\documentclass[11pt]{article}
\usepackage{libertinus}
\usepackage{amsmath,amssymb,amsthm,mathtools}
\usepackage{mathrsfs}
\usepackage[margin=1in]{geometry}
\usepackage{microtype}
\usepackage{xurl}
\usepackage[hidelinks]{hyperref}
\usepackage[square,numbers]{natbib}
\usepackage{etoolbox}

\hypersetup{
    pdftitle={Approximation Families and the Statistical Efficiency of Denoising Score Matching},
    pdfauthor={Brian Miyatake}
}

\makeatletter
\patchcmd{\@maketitle}{\LARGE \@title}
    {\fontsize{16}{19.2}\selectfont\parbox[b]{.9\textwidth}{\centering\@title}}{}{}
\patchcmd{\@maketitle}{\large}{\normalsize}{}{}
\patchcmd{\@maketitle}{\vskip 1em}{\vskip .5em}{}{}
\patchcmd{\@maketitle}{\vskip 1.5em}{\vskip 1em}{}{}
\patchcmd{\abstract}{\quotation}
    {\setlength{\leftmargini}{.075\textwidth}\quotation}{}{}
\patchcmd{\section}{\Large}{\fontsize{13}{15.6}\selectfont}{}{}
\makeatother

\newtheorem{theorem}{Theorem}[section]
\newtheorem{proposition}[theorem]{Proposition}
\newtheorem{lemma}[theorem]{Lemma}
\newtheorem{corollary}[theorem]{Corollary}
\theoremstyle{remark}
\newtheorem{remark}[theorem]{Remark}

\newcommand{\R}{\mathbb{R}}
\newcommand{\T}{\mathbb{T}}
\newcommand{\E}{\mathbb{E}}
\newcommand{\norm}[1]{\left\lVert #1 \right\rVert}
\DeclareMathOperator{\tr}{tr}

\title{Approximation Families and the Statistical Efficiency of Denoising Score Matching}
\author{Brian Miyatake\thanks{Department of Statistics and Data Science, Yale University. Email: \href{mailto:brian.miyatake@yale.edu}{brian.miyatake@yale.edu}}}
\date{}

\begin{document}

\maketitle

\begin{abstract}
We study score matching over finite-dimensional approximation spaces generated by gradients of eigenfunctions of the infinitesimal generator of a forward diffusion. We compare polynomial approximation under one-dimensional Ornstein--Uhlenbeck flow with trigonometric-polynomial approximation under heat flow on the torus. In the Ornstein--Uhlenbeck setting, we show that for every non-Gaussian initial distribution uniformly comparable to the standard Gaussian, the asymptotic score-estimation variance grows exponentially along a subsequence of polynomial degrees. Averaging the score-matching loss under the diffusion semigroup, known as operator-informed score matching (OISM), reduces the variance at each fixed degree but does not eliminate this exponential instability at sufficiently small times. A local minimax lower bound shows that the same high-degree dependence arises when estimating the polynomial projection of the score, rather than being an artifact of OISM. In contrast, under heat flow on the torus, OISM over trigonometric polynomials admits finite-sample control of the score error along the forward diffusion and yields the classical minimax rate for estimating sufficiently smooth densities under expected Kullback--Leibler loss.
\end{abstract}

\section{Introduction and Background}
\label{sec:introduction}

Let $\rho_0$ be an unknown data distribution. Consider a diffusion $(X_t)_{t\geq0}$
on a smooth Riemannian manifold $\mathcal M$ without boundary, with $X_0\sim\rho_0$ and
infinitesimal generator
\begin{equation}
    \mathscr L f=\Delta_{\mathcal M}f+\langle b,\nabla f\rangle,
    \label{eq:generator}
\end{equation}
where $b(x)\in T_x\mathcal M$ for each $x\in\mathcal M$ and $T_x\mathcal M$ 
denotes the tangent space of $\mathcal M$ at $x$.
Here and throughout, $\nabla$ denotes the Riemannian gradient and
$\Delta_{\mathcal M}$ the Laplace--Beltrami operator on $\mathcal M$.
The generator determines the Markov semigroup $P_t=e^{t\mathscr L}$,
and hence the evolution of the forward process. Such forward processes
underlie diffusion-based generative models
\citep{sohldickstein2015deep,ho2020denoising,song2021scorebased}.
Two examples are the one-dimensional Ornstein--Uhlenbeck (OU) process,
corresponding to a variance-preserving (VP) diffusion,
\[
    \mathrm{d}X_t=-X_t\,\mathrm{d}t+\sqrt{2}\,\mathrm{d}B_t,
\]
and heat flow on the flat one-dimensional torus $\T:=\mathbb R/2\pi\mathbb Z$,
the periodic analogue of a variance-exploding (VE) diffusion,
\[
    \mathrm{d}X_t=\sqrt{2}\,\mathrm{d}B_t,
\]
where $\{B_t\}_{t\geq 0}$ is one-dimensional Brownian motion and
the latter equation is understood to be modulo $2\pi$.
Writing $X_t\sim\rho_t$, suppose that $\rho_t$ has density $p_t$ with
respect to Riemannian volume. Its score function is defined as
\[
    s_t(x):=\nabla\log p_t(x)\in T_x\mathcal M.
\]
Then, under suitable regularity, the time-reversed process
$(X_t^{\leftarrow})_{t\in[0,T]}$, initialized from
$X_0^{\leftarrow}\sim\rho_T$, has time-dependent generator
\citep{DBMHTD22,haussmann1986reversal}
\[
    \mathscr L_t^{\leftarrow}f
        =\Delta_{\mathcal M}f
            +\langle -b+2s_{T-t},\nabla f\rangle.
\]
Thus, $\mathscr L$ determines the prescribed forward evolution, while
the reverse dynamics additionally depend on the unknown time-varying
score $s_t$ \citep{chen2023sampling,conforti2025kl}. Estimating this function
from $n$ independent samples of $\rho_0$ is therefore a central statistical problem in
score-based generative modeling.

The score $s_t$ is a time-varying vector field on $\mathcal M$. A natural
idea is to approximate this field by gradients of scalar functions. At a fixed
time $t$, let $\mathcal U$ be a finite-dimensional space of potential
functions and consider the population problem
\begin{equation}
    \inf_{f\in\mathcal U}\frac12
        \int_{\mathcal M}\norm{\nabla f-s_t}^2\,\mathrm{d}\rho_t.
    \label{eq:score-class-risk}
\end{equation}
In this paper, we study growing approximation spaces
\[
    \mathcal U_1\subset\mathcal U_2\subset\cdots.
\]
Increasing $N$ makes the score-approximation class more expressive,
raising the question of how its statistical behavior changes.
Enlarging the approximation space can only improve the best population
approximation to the score, but estimating additional degrees of freedom
from data can increase statistical error. This tradeoff depends strongly
on the choice of approximation family: good approximation properties
alone do not guarantee favorable statistical behavior under score matching.

The formulation in \eqref{eq:generator} suggests choosing approximation spaces adapted to the generator. In particular, if $\mathscr L\psi_j=-\lambda_j\psi_j$, then $P_t\psi_j=e^{-\lambda_j t}\psi_j$, so eigenfunctions of $\mathscr L$ diagonalize the diffusion semigroup. In the two settings considered here, the derivatives and products appearing in the score-matching loss also admit explicit expansions in these eigenfunction families. This makes operator-informed score matching (OISM) \citep{shen2025operator}, which analytically averages the conditional denoising score-matching loss under the forward transition, tractable directly from the initial observations without simulating the forward diffusion; see Section~\ref{subsec:operator-informed}. As we will see, the product structure of these families also plays a central role in their statistical behavior.

This viewpoint associates a score-approximation problem with three
ingredients: the state space $\mathcal M$, the forward generator
$\mathscr L$, and the approximation family $\{\mathcal U_N\}_{N\geq1}$.
We study how score matching depends on the triple
$(\mathcal M,\mathscr L,\{\mathcal U_N\}_{N\geq1})$, with particular
emphasis on the approximation family. For the OU and heat-flow examples
above, we use the following generator-adapted choices.

The first has $\mathcal M=\R$ and
$\mathscr L_{\mathrm{OU}}f=f''-xf'$. The invariant law is the standard
Gaussian, and the generator's eigenfunctions are Hermite polynomials.
Taking the first $N$ nonconstant Hermite polynomials as potentials gives
polynomial score approximation of degree at most $N-1$.

The second has $\mathcal M=\T^d=(\mathbb{R}/2\pi\mathbb{Z})^d$ and $\mathscr L_{\mathrm{heat}}=\Delta$, the Laplace-Beltrami operator. The
invariant law is uniform, and the eigenfunctions are Fourier modes.
Restricting the Fourier frequencies to $|k|_2\leq N$, the approximating
class consists of gradients of trigonometric-polynomial potentials.

Both choices are canonical spectral approximation families for their
respective forward processes. Nevertheless, their statistical behavior
under score matching is sharply different. Polynomial approximation
under OU flow exhibits severe high-degree asymptotic variance growth,
whereas OISM with trigonometric-polynomial approximation under heat flow
admits finite-sample control of the variance contribution at the usual
finite-dimensional scale.

\subsection*{Main results}

We use polynomial instability to refer to exponential growth, along a
subsequence of degrees $N$, of the asymptotic $L^2(\rho_t)$ variance. The
asymptotic variance is defined at fixed $N$, with its behavior then
studied as $N\to\infty$.

\noindent\textbf{Informal Theorem 1 (Polynomial instability under OU flow).}
In the one-dimensional OU setting with polynomial approximation, write
$V_N(t)$ and $V_N^{\mathrm{OI}}(t)$ for the asymptotic
$L^2(\rho_t)$ variances of direct score matching and OISM, respectively.
For every non-Gaussian initial distribution uniformly comparable to the
standard Gaussian and every fixed $t>0$, along possibly different
subsequences of degrees $N$,
\[
    V_N(t)\geq 4^{N-o(N)},
    \qquad
    V_N^{\mathrm{OI}}(t)\geq
        \exp\!\bigl((\log4-4t)N-o(N)\bigr).
\]

\noindent\textbf{Informal Theorem 2 (Fourier variance control under heat flow).}
In the heat-flow setting on $\T^d$ with trigonometric-polynomial
approximation, write $V_N^{\mathrm{OI}}(t)$ for the asymptotic
$L^2(\rho_t)$ variance of OISM. Under uniform upper and lower density
bounds, OISM with frequency cutoff $N$ satisfies
\[
    \int_0^T V_N^{\mathrm{OI}}(t)\,\mathrm{d}t=O(N^d),
    \qquad
    \E\int_0^T
        \norm{\widehat s_{t,N}-\nabla f_{t,N}^\star}_{L^2(\rho_t)}^2\,\mathrm{d}t
        =O\!\left(\frac{N^d}{n}\right).
\]
Here $f_{t,N}^\star$ denotes the population trigonometric-polynomial fit,
and $\widehat s_{t,N}$ denotes the OISM score estimator.

The favorable variance behavior in the Fourier setting also permits the
usual approximation--estimation tradeoff. For an initial density of
smoothness $s$, the integrated squared bias is $O(N^{-2s})$. Combining
this with the $O(N^d/n)$ estimation bound and choosing
$N\asymp n^{1/(2s+d)}$ yields the rate $n^{-2s/(2s+d)}$. The corresponding
approximate reverse diffusion attains this rate under expected
Kullback--Leibler loss, and we show that the rate is minimax optimal.

\paragraph{Polynomial instability: remarks and related work.}
Koehler et~al.\ \citep{koehler2023efficiency} study the statistical
efficiency of score matching relative to maximum likelihood, connecting
its asymptotic covariance to Poincar\'e and isoperimetric properties and
showing that sparse cuts can lead to severe inefficiency. More recently,
L\"utke Schwienhorst, Klein, and Lederer~\citep{LSKL26} study
mixture-weight estimation in a two-component isotropic Gaussian mixture
with known component means. They show that the asymptotic variance of
vanilla score matching grows exponentially with the squared separation
between the components, while diffusion-based denoising score matching
can avoid this unfavorable dependence with a suitable diffusion time.
Qin and Risteski~\citep{qin2024fit} develop generalized score-matching
objectives associated with Markov processes, relate their statistical
efficiency to mixing properties, and obtain polynomial sample-complexity
guarantees for an annealed construction on shared-covariance Gaussian
mixtures. Wenliang and Kanagawa~\citep{wenliang2021blindness} identify
another failure mode of score-based methods: for sufficiently separated
mixtures, Fisher divergence can become insensitive to isolated components
or incorrect mixing proportions.

Our result identifies a distinct instability mechanism. In the preceding
mixture and isoperimetric results, inefficiency is driven by separation,
multimodality, or poor geometric properties of the underlying statistical
model. Here the non-Gaussian law is fixed and uniformly comparable to the
Gaussian, while the asymptotic variance grows exponentially as the
polynomial approximation space expands. The mechanism is instead the
product structure of the polynomial approximation family.

The generator-eigenfunction viewpoint used here is most closely related
to Shen et~al.\ \citep{shen2025operator}, who introduced OISM by using
the diffusion semigroup to average the score-matching objective. They
also report an empirical observation qualitatively consistent with the
high-degree behavior studied here: in their CIFAR-10 experiment under OU
flow, increasing the maximal Hermite order from three to six does not
improve performance, with OISM(6) performing slightly worse than
OISM(3). Because their method additionally fits a neural residual
estimator, this experiment does not isolate the variance mechanism
considered here. Khoo, Oster, and Peng~\citep{khoo2025optimization}
likewise use eigenfunctions of the diffusion generator to obtain
finite-dimensional score representations whose coefficients are
computed from the original observations through linear systems, without
simulating the forward diffusion.

Related work also considers the statistical behavior of score matching
over structured Hermite approximation spaces. Tahmasebi and
Weber~\citep{tahmasebi2026dimension} study invariant score matching on
sets and graphs using growing families of invariant Hermite features.
They derive dimension-free scaling laws governed by the growth of the
feature hierarchy and the decay of the target coefficients, including
regimes with logarithmic rather than polynomial convergence. Chen and van
Leeuwen~\citep{chen2026copula} use finite Hermite expansions and
regularized score-matching linear systems for reduced-order non-Gaussian
density estimation. In contrast to these works, we isolate an explicit
high-degree amplification mechanism for ordinary polynomial score
matching under OU flow and show that it forces exponential asymptotic
variance growth along a subsequence of polynomial degrees.

Several aspects of the polynomial instability result are worth
emphasizing.
\begin{itemize}
    \item \textbf{The approximation family is natural.} Hermite
    polynomials are eigenfunctions of the OU generator, so polynomial
    truncation is a canonical approximation family suggested by the forward diffusion.
    The statistical instability occurs despite this natural matching.

    \item \textbf{The instability is not specific to the Hermite basis.}
    Hermite coordinates are convenient because they diagonalize the
    semigroup, but the fitted score depends on the polynomial
    approximation space itself. Another basis of the same space does
    not remove the phenomenon.

    \item \textbf{The data distribution is not pathological.} The theorem holds for 
    every non-Gaussian initial law whose density ratio with respect to the standard 
    Gaussian is bounded above and below. In particular, it applies to bounded 
    non-Gaussian perturbations that are arbitrarily close to the Gaussian in 
    Kullback–Leibler divergence. Thus the exponential variance growth is not 
    a consequence of the data distribution being far from the invariant Gaussian law.

    \item \textbf{Variance reduction does not remove the pathology.}
    OISM reduces asymptotic variance at every fixed degree, but this improvement does not prevent exponential high-degree growth at sufficiently small diffusion times. Moreover, our local minimax lower bound shows that the same high-degree dependence arises when estimating the polynomial projection of the score, so the obstruction is not merely an artifact of OISM.

    \item \textbf{The result suggests a pessimistic approximation--estimation
    tradeoff.} If analogous exponential variance growth persists when
    degree grows with sample size, usable degrees could grow only
    logarithmically with $n$, suggesting logarithmic rather than
    polynomial convergence rates. We do not prove a growing-degree
    finite-sample lower bound here.
\end{itemize}

\paragraph{Fourier variance control: remarks and related work.}
Optimal or near-optimal rates for score and density estimation have been
established in several diffusion settings. Oko, Akiyama, and
Suzuki~\citep{OAS23} show that score-based diffusion models under
Ornstein--Uhlenbeck smoothing attain nearly minimax density-estimation
rates in total variation over Besov classes. Zhang et~al.\ \citep{ZYLL24}
study Brownian heat flow and obtain nearly minimax total-variation rates
over Sobolev classes using kernel score estimation with early stopping.
Most closely related to our positive result, Dou et~al.\ \citep{dou2024optimal}
establish sharp minimax rates for score estimation over H\"older classes
under Gaussian heat flow and use them to attain the sharp minimax
density-estimation rate in total variation without early stopping or
logarithmic losses. Our result instead considers heat flow on the torus,
where OISM over finite Fourier potential spaces yields an approximate
reverse diffusion attaining the minimax rate under expected
Kullback--Leibler loss,
with the coefficients computed directly from the original observations.

Mardia, Kent, and Laha~\citep{mardia2016score}
develop score matching on compact Riemannian manifolds, including
toroidal models and fixed-dimensional families that can involve
Laplace--Beltrami eigenfunctions. Bonis, Pham~Ngoc, and
Tran~\citep{bonis2026score} study score estimation by constrained and
penalized score-matching empirical risk minimization. In a no-diffusion
torus setting, they recover the classical nonparametric rate associated
with first-derivative estimation; they then analyze ambient OU smoothing
for distributions supported on compact submanifolds and use
Sobolev-constrained score estimators to obtain near-minimax
Wasserstein-$1$ rates for the resulting generative model. In contrast, our
positive result studies OISM along heat flow on the torus over growing
trigonometric-polynomial potential spaces, with integrated-in-time score-error control and
a guarantee on the Kullback--Leibler loss for the generated density.

Shen et~al.\ \citep{shen2025operator} also
illustrate OISM using trigonometric eigenfunctions under Brownian motion
wrapped onto $[-\pi,\pi]$, the one-dimensional periodic analogue of our
heat-flow setting, obtaining accurate score and density estimates from a
finite Fourier family. Chen et~al.\ \citep{chen2023sampling} connect
integrated score estimation along a diffusion to distributional
guarantees for the learned reverse process.

Chewi
et~al.\ \citep{chewi2026ddpm} show that score estimation along Euclidean
diffusions yields near-minimax density-estimation rates under $L^1$ loss
over H\"older classes, using a reduction from score estimation to density
estimation. The smooth
density-estimation exponent is itself classical; see Yang and
Barron~\citep{yang1999information}.

Several features of our positive result are worth noting.
\begin{itemize}
    \item \textbf{Minimax density estimation without forward simulation.}
    We establish the classical minimax rate $n^{-2s/(2s+d)}$ under expected
    Kullback--Leibler loss over the periodic H\"older class considered here.
    Building on the operator-informed formulation of Shen et~al.\ \citep{shen2025operator},
    the Fourier score-matching coefficients can be computed directly from
    the original observations using explicit heat-semigroup formulas. Thus,
    the minimax rate is attained without simulating the forward diffusion.

    \item \textbf{The usual bias--variance tradeoff is explicit.}
    For an initial density of smoothness $s$, Fourier truncation
    contributes $O(N^{-2s})$ integrated squared approximation error,
    while finite-sample estimation contributes $O(N^d/n)$. Their balance
    yields the minimax rate stated above. Our argument has the advantage of being simple and making transparent how the approximation family governs the statistical behavior of score matching.

\end{itemize}

The key distinction lies in how products generated by the quadratic
score-matching objective behave within the approximation family. In the
polynomial setting, the highest-degree term in the fitted score gives rise
to the square $\phi_{N-1}^2$. Its expansion in the Hermite basis contains a
degree-$(2N-2)$ component whose coefficient grows rapidly with $N$.
More precisely, if $\{\phi_j\}_{j\geq0}$ denotes the normalized Hermite
polynomials in $L^2(\pi)$, then the highest-degree component of
$\phi_{N-1}^2$ has squared coefficient
\[
    \binom{2N-2}{N-1}\asymp\frac{4^N}{\sqrt N}.
\]
By contrast, for Fourier functions $\omega_k(x)=e^{i\langle k,x\rangle}$,
\[
    \omega_k\omega_\ell=\omega_{k+\ell}.
\]
Multiplication changes frequency without introducing a coefficient that
grows with the cutoff. Section~\ref{sec:polynomial} makes precise how this
difference in product structure enters the variance of the score-matching
estimator.

Section~\ref{sec:preliminaries} introduces direct score matching and OISM
in a general diffusion setting. Section~\ref{sec:polynomial} specializes
to one-dimensional OU flow and polynomial approximation, derives the
estimators, and establishes the variance-growth results and their
mechanism. Section~\ref{sec:heat} considers heat flow on the torus with
Fourier approximation, derives the OISM estimator, controls
approximation and estimation error, and establishes the minimax
density-estimation rate. Section~\ref{sec:discussion} discusses
implications and open questions.

\paragraph{AI usage.}
This project was motivated by an attempt to understand the 
theoretical guarantees of OISM. The author independently 
investigated the possibility of exponential variance growth, 
identified products of high-degree Hermite polynomials as a 
potential mechanism for instability, and recognized that 
polynomial instability extends beyond OISM to direct score 
matching.

GPT-5.6 Sol was used to explore proof strategies, assist with 
calculations, check proofs, and revise the manuscript. The author 
credits the complex-analytic approximation argument used in the 
OISM variance lower bound under Ornstein--Uhlenbeck flow, as well 
as some calculations in the finite-sample OISM upper bound under 
heat flow to GPT-5.6 Sol. 
The author independently verified the mathematical arguments 
and cited references and takes full responsibility for the 
contents of this work.

\section{Score matching along diffusion processes}
\label{sec:preliminaries}

Let $\mathcal M$ be a smooth Riemannian manifold without
boundary, and let $(X_t)_{t\geq0}$ be a diffusion on $\mathcal M$
with infinitesimal generator $\mathscr L$. Let $(P_t)_{t\geq0}$, with
$P_t=e^{t\mathscr L}$, denote the Markov semigroup generated by
$\mathscr L$. For sufficiently regular functions $h$,
\[
    (P_th)(x)=\E[h(X_t)\mid X_0=x].
\]
Given $X_0\sim\rho_0$, write $X_t\sim\rho_t$. Whenever $\rho_t$
admits a sufficiently regular positive density $p_t$ with respect to
Riemannian volume, we write
\[
    s_t=\nabla\log p_t
\]
for its score. Given a finite-dimensional space $\mathcal U$ of
sufficiently regular potential functions, our goal is to approximate
$s_t$ by gradient fields $\nabla f$, with $f\in\mathcal U$.

We consider two ways of estimating the corresponding score-matching
objective. Direct score matching evaluates the loss along samples evolved
to time $t$. Operator-informed score matching instead averages over the
forward transition using the semigroup $P_t$, allowing the same population
objective to be estimated directly from the initial observations. 

\subsection{Direct score matching}
\label{subsec:score-matching}

Following Hyv\"arinen~\citep{hyvarinen2005} and the Riemannian
formulation of Mardia, Kent, and Laha~\citep{mardia2016score}, for
$f\in\mathcal{U}$, consider the population approximation error
\[
    \mathcal J_t(f)
    := \frac12 \int_{\mathcal M} \|\nabla f-s_t\|^2\,d\rho_t .
\]
When \(\mathcal{U}\) is finite-dimensional, parametrizing \(f\in \mathcal{U}\) by its
coefficients yields a quadratic score-matching objective; see
Hyv\"arinen~\citep{hyvarinen2007extensions}. Up to an additive constant
$C_t$ independent of $f$,
\[
    \begin{aligned}
    \mathcal J_t(f)
        &=\frac12\int_{\mathcal{M}}\norm{\nabla f}^2\,\mathrm{d}\rho_t
            -\int_{\mathcal{M}}\langle\nabla f,s_t\rangle\,\mathrm{d}\rho_t+C_t\\
        &=\frac12\int_{\mathcal{M}}\norm{\nabla f}^2\,\mathrm{d}\rho_t
            -\int_{\mathcal{M}}\langle\nabla f,\nabla p_t\rangle\,\mathrm{d}\operatorname{vol}+C_t\\
        &=\int_{\mathcal{M}}\left(\frac12\norm{\nabla f}^2+\Delta_{\mathcal M} f\right)\,\mathrm{d}\rho_t+C_t.
    \end{aligned}
\]
The final equality follows from integration by parts under the regularity and
integrability assumptions imposed in the concrete settings below. Thus minimizing the
score approximation error is equivalent to minimizing
\[
    L_t(f)=\int_{\mathcal{M}}\ell_f\,\mathrm{d}\rho_t,
    \qquad \ell_f:=\frac12\norm{\nabla f}^2+\Delta_{\mathcal M} f.
\]
Suppose that
\[
    X_0^{(1)},\ldots,X_0^{(n)}\overset{\mathrm{iid}}{\sim}\rho_0,
\]
and, conditionally on these observations, evolve them independently
according to the forward diffusion to obtain
$X_t^{(1)},\ldots,X_t^{(n)}$. Marginally,
\[
    X_t^{(1)},\ldots,X_t^{(n)}\overset{\mathrm{iid}}{\sim}\rho_t.
\]
Direct score matching replaces the population expectation by the empirical objective
\begin{equation}
    \widehat L_t(f)
        =\frac1n\sum_{i=1}^n\ell_f\bigl(X_t^{(i)}\bigr).
    \label{eq:empirical-quadratic}
\end{equation}
In our setting, evaluating~\eqref{eq:empirical-quadratic} therefore entails
evolving the initial observations through the forward diffusion to time $t$.

\subsection{Operator-informed score matching}
\label{subsec:operator-informed}

Operator-informed score matching, introduced by Shen et~al.\ \citep{shen2025operator},
averages the direct score-matching loss over the forward diffusion. By the Markov property,
\[
    (P_t\ell_f)(x)
        =\E\!\left[\ell_f(X_t)\mid X_0=x\right].
\]
Consequently,
\[
    \int_{\mathcal{M}}P_t\ell_f\,\mathrm{d}\rho_0
        =\int_{\mathcal{M}}\ell_f\,\mathrm{d}\rho_t=L_t(f),
\]
so applying the semigroup leaves the population objective unchanged.

For the OU and torus heat kernels used below, let
$p_{t|0}(\,\cdot\mid x)$ denote the forward transition density at $t>0$.
Conditional integration by parts gives
\[
    (P_t\ell_f)(x)
        =\frac12\E\!\left[
            \norm{\nabla f(X_t)
                -\nabla \log p_{t|0}(X_t\mid x)}^2
            \,\middle|\,X_0=x
        \right]+C_{t, x},
\]
where the additive term $C_{t, x}$ is independent of $f$. Thus OISM analytically
averages the denoising score-matching loss under the forward transition
using the semigroup.

Given the same initial observations
\[
    X_0^{(1)},\ldots,X_0^{(n)}\overset{\mathrm{iid}}{\sim}\rho_0,
\]
operator-informed score matching uses the empirical objective
\begin{equation}
    \widehat L_t^{\mathrm{OI}}(f)
        =\frac1n\sum_{i=1}^n(P_t\ell_f)\bigl(X_0^{(i)}\bigr)
        =\frac1n\sum_{i=1}^n
            P_t\!\left(\frac12\norm{\nabla f}^2+\Delta_{\mathcal M} f\right)
                \bigl(X_0^{(i)}\bigr).
    \label{eq:oi-objective}
\end{equation}

The usefulness of this representation depends on whether $P_t\ell_f$
can be evaluated tractably. In general, computing the conditional
expectation defining $P_t\ell_f$ could itself require simulation of the
forward process. The operator-informed formulation becomes particularly
useful when the semigroup acts explicitly on the chosen approximation
family and on the functions appearing in the score-matching loss. In that
case, $P_t\ell_f$ can be computed directly from the initial observation,
without simulating the forward diffusion. In the two settings considered
below, the basis functions diagonalize the corresponding semigroups,
while their products admit explicit expansions in the same eigenfunction
families.

The two procedures therefore have the same population objective and hence
target the same fitted score within $\nabla\mathcal U$. Their empirical
objectives, however, average different random quantities. Direct score
matching uses
\[
    \ell_f(X_t),
\]
whereas operator-informed score matching uses its conditional expectation
\[
    \E[\ell_f(X_t)\mid X_0].
\]
With the two estimation procedures fixed, we now ask how their statistical
behavior changes along a growing family
$\mathcal U_1\subset\mathcal U_2\subset\cdots$.

\section{Instability of polynomial approximation under OU flow}
\label{sec:polynomial}

We first consider the one-dimensional Ornstein--Uhlenbeck process
introduced in Section~\ref{sec:introduction}, with generator
\[
    \mathscr L_{\mathrm{OU}}f=f''-xf',
\]
and invariant distribution $\pi=\mathcal N(0,1)$. Let
$\{\phi_j\}_{j\geq0}$ denote the normalized
Hermite polynomials in $L^2(\pi)$. They satisfy
\[
    \mathscr L_{\mathrm{OU}}\phi_j=-j\phi_j,
    \qquad P_t\phi_j=e^{-jt}\phi_j,
    \qquad \phi_j'=\sqrt j\,\phi_{j-1}.
\]
Thus the Hermite polynomials are eigenfunctions of the forward generator
and provide a natural sequence of approximation spaces. For $N\geq1$, let
\[
    \mathcal U_N:=\operatorname{span}\{\phi_1,\ldots,\phi_N\}.
\]
For $f=\sum_{j=1}^N a_j\phi_j\in\mathcal U_N$,
\[
    P_tf=\sum_{j=1}^N a_je^{-jt}\phi_j\in\mathcal U_N.
\]
Let $\mathcal P_N$ denote the space of real polynomials of degree at most
$N$. The potential space $\mathcal U_N$ is naturally identified with
$\mathcal P_N/\mathbb R$: two polynomial potentials are identified if they
differ only by an additive constant, since constants do not affect their
gradients. In the notation of Section~\ref{sec:preliminaries}, this gives
the triple
\[
    \left(\mathcal M,\mathscr L,\{\mathcal U_N\}_{N\geq1}\right)
    =\left(\R,\mathscr L_{\mathrm{OU}},
        \{\mathcal P_N/\mathbb R\}_{N\geq1}\right).
\]
As the potential ranges over $\mathcal P_N/\mathbb R$, its derivative ranges over
$\mathcal P_{N-1}$, the real polynomials of degree at most $N-1$. At each
fixed time $t>0$, we approximate $s_t$ in this class.

\smallskip
\noindent\textbf{Assumptions.}\quad Let $p_\pi$ denote the standard Gaussian density and write
$r_0:=\mathrm{d}\rho_0/\mathrm{d}\pi$. Throughout this section, assume
\begin{equation}
    0<m\leq r_0\leq M<\infty,
    \qquad r_0\not\equiv1,
    \label{eq:polynomial-density-ratio}
\end{equation}
where the bounds hold $\pi$-almost everywhere. Under OU flow,
$r_t:=p_t/p_\pi=P_tr_0$, so the same bounds hold at every later time.

We use Hermite coordinates because the OU semigroup acts diagonally in this
basis. The fitted score itself depends only on $\mathcal P_{N-1}$.
We emphasize again that whenever the empirical fit is unique, replacing the Hermite polynomials
by another basis of the same polynomial class leaves the fitted score
unchanged. The instability below is therefore a property of polynomial
approximation rather than of the Hermite basis itself.

\subsection{Direct score matching}
\label{subsec:fixed-time-variance-growth}

To parametrize the approximation class, write
\[
    f_c=\sum_{j=1}^N c_j\phi_j,
    \qquad c=(c_1,\ldots,c_N)^\top.
\]
Since $f_c'=\sum_{j=1}^N\sqrt j\,c_j\phi_{j-1}$, these derivatives
span exactly $\mathcal P_{N-1}$. Substituting $f_c$ into the
score-matching loss from Section~\ref{sec:preliminaries} gives the
quadratic function on $\R^N$,
\[
    \ell_{f_c}(x)
        =\frac12 c^\top\mathfrak a(x)c
            -\mathfrak b(x)^\top c,
\]
where
\begin{equation}
    \mathfrak a(x)_{jk}
        =\sqrt{jk}\,\phi_{j-1}(x)\phi_{k-1}(x),
    \qquad
    \mathfrak b(x)_j
        =\begin{cases}
            0,&j=1,\\
            -\sqrt{j(j-1)}\,\phi_{j-2}(x),&j\geq2.
        \end{cases}
    \label{eq:polynomial-coefficients}
\end{equation}
These two quantities have a simple interpretation. The matrix
$\mathfrak a(x)$ is the single-observation contribution to the quadratic
part of the score-matching objective, while $\mathfrak b(x)$ gives the
corresponding linear term. Their empirical averages therefore determine
the normal equations for the fitted coefficients.
Thus, once the polynomial approximation space is fixed, direct score
matching reduces to a finite-dimensional quadratic estimation problem
in the Hermite coefficients $c$.

Averaging this loss over the evolved sample then gives a quadratic empirical
objective in $c$:
\[
    \widehat L_t(f_c)
        =\frac12 c^\top\widehat A_{t,N}c
            -\widehat b_{t,N}^\top c,
\]
where
\[
    \widehat A_{t,N}=\frac1n\sum_{i=1}^n\mathfrak a(X_t^{(i)}),
    \qquad
    \widehat b_{t,N}=\frac1n\sum_{i=1}^n\mathfrak b(X_t^{(i)}).
\]
The data enter this quadratic problem through these empirical coefficients,
which are computed from the evolved observations $X_t^{(i)}$.
Thus, whenever $\widehat A_{t,N}$ is invertible, the empirical
score-matching objective has the unique minimizer
\[
    \widehat c_{t,N}=\widehat A_{t,N}^{-1}\widehat b_{t,N},
\]
and the fitted score is $\widehat s_{t,N}=f_{\widehat c_{t,N}}'$.

To separate approximation from statistical fluctuation, we next identify
the corresponding population fit.
The population analogue is obtained by replacing the sample averages
with expectations. Writing
\[
    A_{t,N}:=\E_{\rho_t}[\mathfrak a(X_t)],
    \qquad b_{t,N}:=\E_{\rho_t}[\mathfrak b(X_t)],
\]
the population score-matching objective is the quadratic function
\[
    L_t(f_c)=\frac12 c^\top A_{t,N}c
        -b_{t,N}^\top c.
\]
The lower bound $r_t\geq m$ implies that $A_{t,N}$ is positive
definite. Therefore, the population objective has the unique minimizer
\[
    c_{t,N}^\star:=A_{t,N}^{-1}b_{t,N},
    \qquad f_{t,N}^\star:=f_{c_{t,N}^\star}.
\]
Its derivative $(f_{t,N}^\star)'$ is the $L^2(\rho_t)$-orthogonal
projection of $s_t$ onto $\mathcal P_{N-1}$.
This is the best score approximation available in the chosen polynomial
class. To describe fluctuations of the empirical fit around this population
target, define the single-observation residual in the population normal
equations by
\[
    \psi_{t,N}(x):=\mathfrak b(x)-\mathfrak a(x)c_{t,N}^\star.
\]
The population normal equations give
$\E_{\rho_t}[\psi_{t,N}(X_t)]=0$, so its covariance is
\[
    \Sigma_{t,N}:=\E_{\rho_t}\!\left[
        \psi_{t,N}(X_t)\psi_{t,N}(X_t)^\top\right].
\]
Thus $\psi_{t,N}(x)$ measures the single-observation fluctuation of the
normal equations around the population optimum. Its covariance
$\Sigma_{t,N}$ determines the asymptotic fluctuations of the fitted
coefficients and, consequently, the asymptotic variance of the fitted
score.
Appendix~\ref{subsec:score-matching-asymptotics} shows that the
asymptotic $L^2(\rho_t)$ variance of the fitted score is
\[
    V_N(t):=\tr\!\left(A_{t,N}^{-1}\Sigma_{t,N}\right).
\]
Specifically, $V_N(t)$ is the mean of the limiting quadratic form
governing $n\norm{\widehat s_{t,N}-(f_{t,N}^\star)'}_{L^2(\rho_t)}^2$.

We now study how this fixed-$N$ asymptotic variance behaves as the
polynomial cutoff $N$ increases.

\begin{samepage}
\begin{theorem}[Fixed-time subsequential variance growth]
\label{thm:fixed-law-variance-growth}
Let $\rho_0$ satisfy~\eqref{eq:polynomial-density-ratio}.
For every fixed $t>0$,
\begin{equation}
    \limsup_{N\to\infty}\frac1N\log V_N(t)\geq\log4.
    \label{eq:fixed-law-variance-growth}
\end{equation}
\end{theorem}
\end{samepage}
Thus, even at a single positive time, enlarging the polynomial
approximation class can produce exponentially increasing asymptotic
variance along a subsequence. This holds for every non-Gaussian initial law
satisfying the uniform comparison assumption. The subsequence may depend on $t$.

This is a statement about asymptotic variance: first $n\to\infty$ at
fixed $N$, and then the resulting variance is studied as $N\to\infty$.
It is not a finite-sample lower bound for a growing cutoff $N=N(n)$.

\subsection{Operator-informed score matching}
\label{sec:oi-polynomial}

Operator-informed score matching replaces each direct score-matching
contribution by its conditional expectation given the initial observation.
Write $\mathfrak a_t:=P_t\mathfrak a$ and
$\mathfrak b_t:=P_t\mathfrak b$ for the semigroup-averaged
single-observation coefficients.
Since the Hermite polynomials diagonalize the OU semigroup, these
conditional expectations can be evaluated explicitly. Together with the
Hermite product formula, this gives, for $1\leq j,k\leq N$,
\[
    (\mathfrak a_t(x))_{jk}
        =\sqrt{jk}\sum_{r=0}^{\min(j-1,k-1)}
            r!\binom{j-1}{r}\binom{k-1}{r}
            \sqrt{\frac{(j+k-2-2r)!}{(j-1)!(k-1)!}}\,
            e^{-(j+k-2-2r)t}\phi_{j+k-2-2r}(x),
\]
and
\[
    (\mathfrak b_t(x))_j
        =\begin{cases}
            0,&j=1,\\
            -\sqrt{j(j-1)}\,e^{-(j-2)t}\phi_{j-2}(x),&j\geq2.
        \end{cases}
\]
The effect of semigroup averaging is therefore explicit in Hermite
coordinates: each Hermite component appearing in the direct score-matching
coefficients is damped according to its degree. This damping will be the
source of the variance reduction below.

By averaging these coefficients over the original observations, OISM
forms the empirical quadratic objective
\[
    \widehat L_t^{\mathrm{OI}}(f_c)
        =\frac12 c^\top\widehat A_{t,N}^{\mathrm{OI}}c
            -(\widehat b_{t,N}^{\mathrm{OI}})^\top c,
\]
where
\[
    \widehat A_{t,N}^{\mathrm{OI}}
        =\frac1n\sum_{i=1}^n\mathfrak a_t(X_0^{(i)}),
    \qquad
    \widehat b_{t,N}^{\mathrm{OI}}
        =\frac1n\sum_{i=1}^n\mathfrak b_t(X_0^{(i)}).
\]
In particular, these coefficients are computed from the original
observations $X_0^{(i)}$, so no simulation of $X_t^{(i)}$ is required.
Whenever $\widehat A_{t,N}^{\mathrm{OI}}$ is invertible, the unique
minimizer of this empirical objective is
\[
    \widehat c_{t,N}^{\mathrm{OI}}
        =(\widehat A_{t,N}^{\mathrm{OI}})^{-1}\widehat b_{t,N}^{\mathrm{OI}},
\]
with fitted score
$\widehat s_{t,N}^{\mathrm{OI}}=f_{\widehat c_{t,N}^{\mathrm{OI}}}'$.

Semigroup averaging preserves the population objective, so the population
minimizer remains $c_{t,N}^\star$ and $f_{t,N}^\star$ from
Section~\ref{subsec:fixed-time-variance-growth}. Indeed,
\[
    \E_{\rho_0}[\mathfrak a_t(X_0)]=A_{t,N},
    \qquad
    \E_{\rho_0}[\mathfrak b_t(X_0)]=b_{t,N}.
\]
Therefore, the two procedures share the same population target but average different
random quantities. Thus their statistical comparison reduces to comparing
the fluctuations of the empirical normal equations around the common
population minimizer $c_{t,N}^\star$.
To quantify the OISM fluctuations, define
\[
    \psi_{t,N}^{\mathrm{OI}}(x)
        :=\mathfrak b_t(x)-\mathfrak a_t(x)c_{t,N}^\star,
    \qquad
    \Sigma_{t,N}^{\mathrm{OI}}
        :=\E_{\rho_0}\!\left[
            \psi_{t,N}^{\mathrm{OI}}(X_0)
            \psi_{t,N}^{\mathrm{OI}}(X_0)^\top\right],
\]
and
\[
    V_N^{\mathrm{OI}}(t)
        :=\tr\!\left(A_{t,N}^{-1}\Sigma_{t,N}^{\mathrm{OI}}\right).
\]
Because $\psi_{t,N}^{\mathrm{OI}}(X_0)$ is the conditional expectation
of $\psi_{t,N}(X_t)$ given $X_0$, the conditional covariance identity gives
\[
    V_N^{\mathrm{OI}}(t)\leq V_N(t).
\]
OISM therefore reduces asymptotic variance at every fixed degree. The
next result shows that this reduction need not prevent exponential growth
as the degree increases.

\begin{samepage}
\begin{theorem}[Operator-informed asymptotic variance lower bound]
\label{thm:oi-polynomial-variance-growth}
Assume~\eqref{eq:polynomial-density-ratio}.
For every fixed $t>0$,
\begin{equation}
    \limsup_{N\to\infty}\frac1N\log V_N^{\mathrm{OI}}(t)
        \geq\log4-4t.
    \label{eq:oi-pointwise-variance-growth}
\end{equation}
\end{theorem}
\end{samepage}
The theorem should be read together with the fixed-degree variance
reduction $V_N^{\mathrm{OI}}(t)\leq V_N(t)$. Operator averaging reduces
asymptotic variance at every fixed degree, but for $t<(\log4)/4$ its
damping does not offset the high-degree product amplification.
For $t\geq(\log4)/4$, the fixed-time lower bound no longer implies
exponential growth.

\begin{remark}[Higher-dimensional embedding]
The one-dimensional instability embeds into OU flow in every ambient
dimension. Indeed, let
\[
    \rho_0=\mu_0\otimes\pi_{d-1},
\]
where $\mu_0$ satisfies the assumptions above and $\pi_{d-1}$ is the
standard Gaussian law on $\mathbb R^{d-1}$. Then
\[
    \rho_t=\mu_t\otimes\pi_{d-1}
\]
under $d$-dimensional OU flow. For the usual total-degree multivariate
Hermite approximation, the population score-matching quadratic form is
block diagonal between Hermite potentials depending only on the first
coordinate and those involving any of the remaining coordinates. The
first block coincides exactly with the one-dimensional problem considered
above. Consequently, in both direct score matching and OISM, the full
asymptotic variance contains the corresponding one-dimensional asymptotic
variance as a nonnegative contribution, so the variance lower bounds of
Theorems~\ref{thm:fixed-law-variance-growth}
and~\ref{thm:oi-polynomial-variance-growth} persist for this class of $d$-dimensional product
distributions. We do not claim the analogous universal statement for
arbitrary non-Gaussian distributions on $\mathbb R^d$.
\end{remark}

\subsection{Mechanism behind the variance growth}
\label{subsec:ou-instability-mechanism}

The preceding results raise a natural question: why can enlarging a
natural approximation class make score matching statistically inefficient?
There are two competing effects. Adding a polynomial degree improves the
population approximation to the score, but the corresponding coefficient
also enters products of polynomial features in the empirical score-matching
objective. The proof compares these two effects.

To measure the approximation improvement, let
\[
    B_N^{\mathrm{OU}}(t):=\inf_{w\in\mathcal P_{N-1}}
        \norm{s_t-w}_{L^2(\rho_t)}^2
\]
denote the best squared approximation error at degree $N-1$, and for
$N\geq2$ define
\[
    \Delta_N(t):=B_{N-1}^{\mathrm{OU}}(t)-B_N^{\mathrm{OU}}(t)
\]
as the nonnegative improvement obtained by adding the degree-$(N-1)$
component. Throughout Section~\ref{sec:polynomial}
and Appendix~\ref{app:polynomial-proofs}, we use the convention
$\log 0=-\infty$ when taking logarithms of nonnegative quantities. Write
$\alpha_N(t):=(c_{t,N}^\star)_N$ for the highest coefficient of the
fitted potential. Since $\phi_N'=\sqrt N\,\phi_{N-1}$, the corresponding
fitted-score coefficient is $\sqrt N\,\alpha_N(t)$.

\begin{lemma}[Approximation increment]
\label{lem:bias-increments}
Assume~\eqref{eq:polynomial-density-ratio}.
For every $N\geq2$ and $t>0$,
\begin{equation}
    \frac{\Delta_N(t)}{M}\leq N\alpha_N(t)^2
        \leq\frac{\Delta_N(t)}{m}.
    \label{eq:bias-increments}
\end{equation}
\end{lemma}
\begin{proof}
The fitted scores at cutoffs $N$ and $N-1$ are the $L^2(\rho_t)$
orthogonal projections onto $\mathcal P_{N-1}$ and
$\mathcal P_{N-2}$. Their difference is
\[
    \sqrt N\,\alpha_N(t)
        \left(\phi_{N-1}
            -\operatorname{proj}^{L^2(\rho_t)}_{\mathcal P_{N-2}}
                \phi_{N-1}\right).
\]
By $m\leq r_t\leq M$,
\[
    \begin{aligned}
    m&\leq m\inf_{q\in\mathcal P_{N-2}}
        \norm{\phi_{N-1}-q}_{L^2(\pi)}^2\\
     &\leq\norm{\phi_{N-1}
        -\operatorname{proj}^{L^2(\rho_t)}_{\mathcal P_{N-2}}
            \phi_{N-1}}_{L^2(\rho_t)}^2\\
     &\leq\norm{\phi_{N-1}}_{L^2(\rho_t)}^2\\&\leq M.
    \end{aligned}
\]
The infimum equals $1$ by Hermite orthogonality, while the upper bound
uses $0\in\mathcal P_{N-2}$ as a competitor.
Properties of the projection give $\Delta_N(t)$ as the squared norm of
this difference, proving~\eqref{eq:bias-increments}.
\end{proof}

If estimation variance depended only on this new coefficient, there would
be no reason to expect instability. The difficulty is that score matching
is quadratic in the polynomial features. In particular, the highest
fitted coefficient enters through the product $\phi_{N-1}^2$:
\[
    \phi_{N-1}^2
        =\sqrt{\binom{2N-2}{N-1}}\,\phi_{2N-2}
            +\text{lower-degree terms},
    \qquad \binom{2N-2}{N-1}\asymp\frac{4^N}{\sqrt N}.
\]
Thus a coefficient introduced to improve approximation at degree $N-1$
creates a much larger degree-$(2N-2)$ component in the score-matching
residual.

\begin{proposition}[Residual amplification]
\label{prop:ou-residual-amplification}
Assume~\eqref{eq:polynomial-density-ratio}.
For every $N\geq3$ and $t>0$,
\begin{equation}
    V_N(t)\gtrsim_{m,M}\Delta_N(t)
        \binom{2N-2}{N-1},
    \label{eq:direct-variance-bias-increment}
\end{equation}
while
\begin{equation}
    V_N^{\mathrm{OI}}(t)
        \gtrsim_{m,M}\Delta_N(t)e^{-4(N-1)t}
            \binom{2N-2}{N-1}.
    \label{eq:oi-variance-bias-increment}
\end{equation}
\end{proposition}
The direct bound compares the approximation improvement with product
amplification:
\[
    V_N(t)\gtrsim_{m,M}
        \underbrace{\Delta_N(t)}_{\text{approximation improvement}}
        \underbrace{\binom{2N-2}{N-1}}_{\text{product amplification}}.
\]
For OISM, the same comparison contains the additional factor
$e^{-4(N-1)t}$, which records the damping of the high-degree component
by the OU semigroup.

\begin{proof}
The degree-$(2N-2)$ coefficient in the $N$th direct residual coordinate is
\[
    -N\alpha_N(t)\sqrt{\binom{2N-2}{N-1}},
\]
and no other term in that coordinate has the same degree. Hermite
orthogonality and $m\leq r_t\leq M$ bound the variance of this residual
coordinate from below. The comparison of that coordinate's variance with
$V_N(t)$, proved in Appendix~\ref{app:polynomial-proofs} by
Cauchy--Schwarz in the $A_{t,N}$-inner product, and
Lemma~\ref{lem:bias-increments} give the direct bound. For OISM, the
same component is multiplied by $e^{-(2N-2)t}$, so its squared
contribution acquires the factor $e^{-4(N-1)t}$. Appendix~\ref{app:polynomial-proofs}
gives the remaining algebra.
\end{proof}

The only way for the exponential product amplification above to be
cancelled would be for $\Delta_N(t)$ itself to decay exponentially fast
with $N$. The final ingredient rules this out.

\begin{lemma}[Non-exponential decay of approximation increments]
\label{lem:non-exponential-increments}
Under~\eqref{eq:polynomial-density-ratio}, for every fixed $t>0$,
\begin{equation}
    \limsup_{N\to\infty}\frac1N\log\Delta_N(t)=0.
    \label{eq:non-exponential-increments}
\end{equation}
\end{lemma}
The proof is deferred to Appendix~\ref{app:polynomial-proofs}. Exponential
decay of the approximation increments would imply exponential decay of the
Hermite coefficients of $s_t$, forcing $s_t$ to admit an entire extension.
Mehler's formula and Hadamard factorization show that this is possible
under the standing assumption only in the stationary Gaussian case
$r_0\equiv1$.

The central-binomial estimate, Proposition~\ref{prop:ou-residual-amplification},
and Lemma~\ref{lem:non-exponential-increments} give
\[
    \limsup_{N\to\infty}\frac1N\log V_N(t)\geq\log4,
    \qquad
    \limsup_{N\to\infty}\frac1N\log V_N^{\mathrm{OI}}(t)
        \geq\log4-4t.
\]
These prove the fixed-time results.

\subsection{Local minimax lower bound}
\label{subsec:ou-local-minimax}

The preceding results concern the asymptotic variance of score-matching
estimators. We next show that the same quantity also controls a local minimax
lower bound for estimating the polynomial projection of the score. Thus the
exponential behavior in Theorem~\ref{thm:oi-polynomial-variance-growth} is not
merely a feature of OISM.

\begin{theorem}[Local minimax lower bound for polynomial score projection]
\label{thm:ou-local-minimax}
Assume~\eqref{eq:polynomial-density-ratio}, and suppose in addition that for some $\eta>0$,
\[
    m+\eta\leq r_0\leq M-\eta
    \qquad \pi\text{-a.e.}
\]
Fix $t>0$ and $N\geq1$, and define
\[
    \Gamma_{t,N}
    :=A_{t,N}^{-1/2}\Sigma_{t,N}^{\mathrm{OI}}A_{t,N}^{-1/2},
    \qquad
    q:=\operatorname{rank}(\Gamma_{t,N}).
\]
Then, for all sufficiently large $n$, there is a family
\[
    \{\rho_{0,\theta,n}:\theta\in\{-1,1\}^q\}
\]
of initial distributions satisfying~\eqref{eq:polynomial-density-ratio} such
that
\[
    \max_{\theta\in\{-1,1\}^q}
    \left\|
        \frac{\mathrm{d}\rho_{0,\theta,n}}{\mathrm{d}\rho_0}-1
    \right\|_\infty
    =O(n^{-1/2}),
\]
where the implied constant may depend on the fixed quantities
$t,N,\rho_0$ and the comparison parameters, but not on $n$ or $\theta$, and
\[
    \begin{aligned}
    &\inf_{\widehat s}
    \sup_{\theta\in\{-1,1\}^q}
    \E_{\rho_{0,\theta,n}^{\otimes n}}
    \left[
        \left\|
            \widehat s-
            \operatorname{proj}_{\mathcal P_{N-1}}^{L^2(\rho_{t,\theta,n})}
            s_{t,\theta,n}
        \right\|_{L^2(\rho_{t,\theta,n})}^2
    \right]\geq
        c\,\frac{V_N^{\mathrm{OI}}(t)}{n},
    \end{aligned}
\]
where $c>0$ is a universal numerical constant, $\rho_{t,\theta,n}$ is the law
obtained by evolving $\rho_{0,\theta,n}$ under OU flow for time $t$, and
$s_{t,\theta,n}$ is its score. The infimum is over all measurable estimators
$\widehat s$ of the $L^2(\rho_{t,\theta,n})$-orthogonal projection of
$s_{t,\theta,n}$ onto $\mathcal P_{N-1}$, the space of real polynomials
of degree at most $N-1$, based on
\[
    X_0^{(1)},\ldots,X_0^{(n)}
    \overset{\mathrm{iid}}{\sim}\rho_{0,\theta,n}.
\]
\end{theorem}

The proof is given in Appendix~\ref{subsec:ou-local-minimax-proof}.

The theorem shows that the unfavorable high-degree behavior of OISM is not 
an artifact of the estimator. For each fixed $N$,
Theorem~\ref{thm:ou-local-minimax} gives a local minimax lower bound of order
$V_N^{\mathrm{OI}}(t)/n$ for estimating the polynomial projection of the
score. Thus, no estimator based on the initial observations can uniformly
remove this dependence in a neighborhood of $\rho_0$. Studying the
coefficient $V_N^{\mathrm{OI}}(t)$ subsequently as $N\to\infty$,
Theorem~\ref{thm:oi-polynomial-variance-growth}
implies exponential growth along a subsequence of degrees when
$t<(\log4)/4$. This is an iterated fixed-$N$ local asymptotic statement,
not a finite-sample lower bound uniform over a growing cutoff $N=N(n)$.

This also clarifies the role of regularization. In the fixed-$N$ local
asymptotic regime, Theorem~\ref{thm:ou-local-minimax} applies to all
measurable estimators of the polynomial projection of the score, including
regularized or shrinkage-based procedures. Thus regularization cannot
uniformly eliminate the dependence on $V_N^{\mathrm{OI}}(t)$ while
retaining the same projected-score target. It can instead reduce the
contribution of high-degree coefficients by shrinking them, but this
introduces bias relative to that target.
Proposition~\ref{prop:ou-residual-amplification} and
Lemma~\ref{lem:bias-increments} make the associated tradeoff explicit: the
highest-degree fitted coefficient is responsible both for the amplified
residual contribution and for the approximation improvement $\Delta_N(t)$
obtained by adding that degree.

\section{Minimax estimation via trigonometric polynomials under heat flow}
\label{sec:heat}
In the notation of Section~\ref{sec:preliminaries}, we take
$\mathcal M=\T^d$ and $\mathscr L=\Delta$. Let $\mathcal T_N$ denote the real
trigonometric-polynomial potential space with nonzero frequencies
$0<|k|_2\leq N$,
defined precisely in Section~\ref{subsec:heat-score-error}. The resulting
triple is
\[
    \left(\mathcal M,\mathscr L,\{\mathcal U_N\}_{N\geq1}\right)
    =\left(\T^d,\Delta,\{\mathcal T_N\}_{N\geq1}\right).
\]

Section~\ref{sec:polynomial} showed that the statistical behavior of score
matching can depend strongly on the product structure of the approximation
family. Under polynomial approximation, products of high-degree Hermite
features introduce coefficients that grow rapidly with the degree. For
$k\in\mathbb Z^d$, write $\omega_k(x):=e^{i\langle k,x\rangle}$. These
Fourier modes behave differently under multiplication:
\[
    \omega_k\omega_\ell=\omega_{k+\ell}.
\]
Multiplication changes the frequency without introducing a coefficient that
grows with the cutoff. We show that, under heat flow, this structure leads to
the usual approximation--estimation tradeoff.
Our heat-flow analysis focuses on the operator-informed estimator; the
statistical behavior of direct score matching under Fourier approximation
is not analyzed here.

\smallskip
\noindent\textbf{Assumptions.}\quad Throughout this section, $d$ is fixed, $\nu$ is the normalized Riemannian
volume measure on $\T^d$, equivalently the uniform probability measure,
and $\rho_0$ has density $p_0$ with respect to $\nu$. We assume
\[
    0<m<1<M<\infty,\qquad m\leq p_0\leq M.
\]
Heat flow preserves these bounds. For an integer $s\geq1$,
$C^{s-1,1}(\T^d)$ consists of functions with continuous derivatives through
order $s-1$ whose derivatives of order $s-1$ are Lipschitz. For the
$2\pi$-periodic lift of $f$, write
\[
    \operatorname{Lip}(f):=
        \sup_{\substack{x,y\in\R^d\\x\neq y}}
            \frac{|f(x)-f(y)|}{|x-y|_2}.
\]
Fix $L>0$ and define the periodic H\"older density class
\begin{equation}
    \mathcal P_s(m,M,L):=
    \left\{p\in C^{s-1,1}(\T^d):\begin{array}{l}
        \displaystyle\int_{\T^d}p\,\mathrm{d}\nu=1,\quad
        m\leq p\leq M,\\[2pt]
        \displaystyle\max_{|\beta|=s-1}
            \operatorname{Lip}(D^\beta p)\leq L
    \end{array}\right\}.
    \label{eq:heat-density-class}
\end{equation}
Here $D^\beta p$ denotes the classical derivative for $|\beta|\leq s-1$.
When $s=1$, the last condition is $\operatorname{Lip}(p)\leq L$.
For $p_0\in\mathcal P_s(m,M,L)$, the initial score
$s_0=\nabla\log p_0$ is interpreted as a weak gradient when $s=1$.
At positive times, heat smoothing gives a classical score. In statements
assuming only the two-sided density bounds, score quantities are understood
for $t>0$, with integrals starting at zero interpreted over $(0,T]$.
Lower-order derivative bounds follow from periodicity and are proved in
Lemma~\ref{lem:heat-holder-derivatives} of Appendix~\ref{app:heat-proofs}.
A related formulation using a highest-order H\"older condition and separate
two-sided density bounds appears in
Chewi et~al.\ \citep[Definition~7]{chewi2026ddpm} for a one-dimensional interval.
The variance analysis in Section~\ref{subsec:heat-variance-estimation} uses
only the two-sided density bounds, while the approximation and minimax
results use the full class $\mathcal P_s(m,M,L)$.

\subsection{Operator-informed score matching}
\label{subsec:heat-score-error}

The infinitesimal generator of heat flow on $\T^d$ is
$\mathscr L_{\mathrm{heat}}=\Delta$, and its invariant probability
measure is the uniform measure $\nu$.
Identifying the torus with $(-\pi,\pi]^d$ with opposite faces identified,
we use $\mathrm{d}\nu(x)=(2\pi)^{-d}\,\mathrm{d}x$. The density at time $t$ is
$p_t=P_tp_0$, and its score is $s_t=\nabla\log p_t$.
For $k\in\mathbb Z^d$, write
$|k|_2=(\sum_{j=1}^d k_j^2)^{1/2}$.
For a frequency cutoff $N\geq1$, let
\begin{equation}
    \Lambda_N:=\{k\in\mathbb Z^d:0<|k|_2\leq N\},
    \qquad |\Lambda_N|\asymp_d N^d.
    \label{eq:heat-frequency-set}
\end{equation}
The lattice-counting argument is given in
Appendix~\ref{subsec:heat-population-fit}. We take $\mathcal U_N=\mathcal T_N$
to be the real trigonometric-polynomial potentials with frequencies in $\Lambda_N$
and approximate the score in
$\mathcal F_N=\{\nabla f:f\in\mathcal U_N\}$.

For computation it is convenient to use these complex Fourier coordinates.
For $c\in\mathbb C^{\Lambda_N}$, write
\[
    f_c(x):=\sum_{k\in\Lambda_N}c_k\omega_k(x).
\]
For arbitrary complex coefficients, extend the real-valued loss from
Section~\ref{subsec:score-matching} by defining
\[
    \ell_{f_c}(x)
        :=\frac12\overline{\nabla f_c(x)}\cdot\nabla f_c(x)
            +\operatorname{Re}\Delta f_c(x).
\]
This agrees with the original score-matching loss when $f_c$ is
real-valued, equivalently when $c_{-k}=\overline{c_k}$.
For $k\in\Lambda_N$,
\[
    \mathscr L_{\mathrm{heat}}\omega_k=-|k|_2^2\omega_k,
    \qquad P_t\omega_k=e^{-t|k|_2^2}\omega_k.
\]
Thus the Fourier modes are eigenfunctions of the heat generator and
provide the natural approximation family in this setting.
A coefficient vector satisfying $c_{-k}=\overline{c_k}$ represents a
real-valued trigonometric potential. Appendix~\ref{subsec:heat-complex-proof}
shows that complexifying the quadratic problem does not change its
population minimum and that uniqueness forces the population and empirical
minimizers to be conjugate symmetric.

In these coordinates, let $\mathfrak a$ and $\mathfrak b$ denote the direct
score-matching quadratic coefficients, defined for $k,\ell\in\Lambda_N$ by
\[
    \mathfrak a(x)_{k\ell}
        :=\nabla\overline{\omega_k}(x)\cdot\nabla\omega_\ell(x),
    \qquad
    \mathfrak b(x)_k:=-\Delta\overline{\omega_k}(x),
\]
Since our heat-flow analysis concerns only OISM, we pass directly from
the direct score-matching coefficients $(\mathfrak a,\mathfrak b)$ to their
semigroup averages, rather than introducing a separate direct empirical
estimator. Write
$\mathfrak a_t:=P_t\mathfrak a$ and $\mathfrak b_t:=P_t\mathfrak b$ for
their semigroup averages.
Specializing the OISM construction of Section~\ref{subsec:operator-informed},
the single-observation coefficients are, for $k,\ell\in\Lambda_N$ and
$t\geq0$,
\begin{equation}
    \begin{aligned}
    (\mathfrak a_t(x))_{k\ell}
        &:=P_t\!\left(\overline{\nabla\omega_k}\cdot
            \nabla\omega_\ell\right)(x)
          =(k\cdot\ell)e^{-t|\ell-k|_2^2}\omega_{\ell-k}(x),\\
    (\mathfrak b_t(x))_k
        &:=-P_t\Delta\overline{\omega_k}(x)
          =|k|_2^2e^{-t|k|_2^2}\omega_{-k}(x).
    \end{aligned}
    \label{eq:heat-fourier-coefficients}
\end{equation}
Thus the semigroup action on the quadratic coefficients is explicit in
Fourier coordinates, while multiplication changes frequency without
introducing the coefficient amplification seen for Hermite products.
These formulas follow from the Fourier product rule and the diagonal
action of the heat semigroup. Appendix~\ref{subsec:heat-complex-proof}
gives the derivation. With $\dagger$ denoting conjugate transpose, the
averaged loss is
\[
    (P_t\ell_{f_c})(x)
        =\frac12 c^\dagger\mathfrak a_t(x)c
            -\operatorname{Re}(\mathfrak b_t(x)^\dagger c).
\]

The population objective averages these coefficients under $\rho_0$:
$A_{t,N}:=\E_{\rho_0}\mathfrak a_t(X_0)$ and
$b_{t,N}:=\E_{\rho_0}\mathfrak b_t(X_0)$.
By the semigroup identity, $A_{t,N}=\E_{\rho_t}\mathfrak a(X_t)$ and
$b_{t,N}=\E_{\rho_t}\mathfrak b(X_t)$.
Since $p_t\geq m$, $A_{t,N}$ is positive definite. Thus the unique
minimizer of the population OISM objective, which is the same population
score-matching objective as in Section~\ref{subsec:operator-informed}, is
\[
    c_{t,N}^\star:=A_{t,N}^{-1}b_{t,N},
    \qquad f_{t,N}^\star:=f_{c_{t,N}^\star}.
\]
From the same initial sample at every time, we estimate these
expectations by
\[
    \widehat A_{t,N}^{\mathrm{OI}}
        :=\frac1n\sum_{i=1}^n\mathfrak a_t(X_0^{(i)}),
    \qquad
    \widehat b_{t,N}^{\mathrm{OI}}
        :=\frac1n\sum_{i=1}^n\mathfrak b_t(X_0^{(i)}).
\]
Thus the OISM fit at every time $t$ is computed from the same original
observations $X_0^{(i)}$, without simulating the forward heat flow.
Whenever the empirical matrix is invertible, the fitted coefficients
and score are
\[
    \widehat c_{t,N}^{\mathrm{OI}}
        =(\widehat A_{t,N}^{\mathrm{OI}})^{-1}
            \widehat b_{t,N}^{\mathrm{OI}},
    \qquad
    \widehat s_{t,N}^{\mathrm{OI}}
        =\sum_{k\in\Lambda_N}ik\widehat c_{t,N,k}^{\mathrm{OI}}\omega_k.
\]
By Appendix~\ref{subsec:heat-complex-proof}, uniqueness of this minimizer
implies $\widehat c_{t,N,-k}^{\mathrm{OI}}
=\overline{\widehat c_{t,N,k}^{\mathrm{OI}}}$, so the fitted potential
and score are real-valued.

For the finite-sample analysis, it is useful to compare the empirical
quadratic form to a deterministic reference scale. Under the uniform
measure $\nu$, Parseval's identity identifies the scale
$K_N:=\operatorname{diag}(|k|_2^2:k\in\Lambda_N)$ through
\[
    c^\dagger K_Nc
        =\int_{\T^d}|\nabla f_c|^2\,\mathrm{d}\nu,
    \qquad
    c^\dagger A_{0,N}c
        =\int_{\T^d}|\nabla f_c|^2p_0\,\mathrm{d}\nu.
\]
Thus $K_N$ provides the natural deterministic scale for the quadratic
form in Fourier coordinates. Since $m\leq p_0\leq M$, the population matrix
at time zero satisfies $mK_N\preceq A_{0,N}\preceq MK_N$. We use the event
\begin{equation}
    \mathcal C_N:=\left\{
        \widehat A_{0,N}^{\mathrm{OI}}\succeq\frac m2K_N
    \right\}.
    \label{eq:heat-conditioning-event}
\end{equation}
On this event, the empirical quadratic form retains a fixed fraction
of the population lower bound. Section~\ref{subsec:heat-variance-estimation}
shows that $\mathcal C_N$ has high probability and that this lower bound
propagates to every later time. Set
$\widehat s_{t,N}=\widehat s_{t,N}^{\mathrm{OI}}$ on $\mathcal C_N$ and
$\widehat s_{t,N}=0$ on $\mathcal C_N^c$. This defines the estimator on
every dataset and permits an unconditional expectation bound.

For a time-dependent score estimator $\widehat s$, define
\begin{equation}
    \mathcal R_T(\widehat s)
        :=\int_0^T\norm{\widehat s_t-s_t}_{L^2(\rho_t)}^2\,\mathrm{d}t.
    \label{eq:integrated-score-risk}
\end{equation}
This integrated score error is the quantity controlled below and later
transferred to Kullback--Leibler loss in Section~\ref{subsec:heat-density-kl}.
Here and below, $\mathbb P_{p_0}$ and $\E_{p_0}$ denote probability and
expectation over the training sample drawn from $p_0\,\mathrm{d}\nu$.

We now quantify the integrated score error of this estimator as the Fourier
cutoff grows. Increasing $N$ improves score approximation but increases
the number of coefficients fitted from data. The next result makes this
approximation--estimation tradeoff precise, with terms at scales
$N^{-2s}$ and $N^d/n$ and an exponentially small conditioning term.

\begin{theorem}[Finite-sample Fourier OISM bound]
\label{thm:heat-expected-score-risk}
For every $N\geq1$, $n\geq1$, and $T>0$,
\begin{equation}
    \sup_{p_0\in\mathcal P_s(m,M,L)}
        \E_{p_0}\mathcal R_T(\widehat s_{\cdot,N})
        \leq C\left(N^{-2s}+\frac{|\Lambda_N|}{n}
            +|\Lambda_N|e^{-cn/|\Lambda_N|}\right),
    \label{eq:heat-expected-score-bound}
\end{equation}
where $C=C(d,s,m,M,L)$ and $c=c(m)>0$ are independent of $\, T,N,n$.
In particular, if $N_n\asymp n^{1/(2s+d)}$, then, for all sufficiently
large $n$,
\begin{equation}
    \sup_{p_0\in\mathcal P_s(m,M,L)}
        \E_{p_0}\mathcal R_T(\widehat s_{\cdot,N_n})
        \leq Cn^{-2s/(2s+d)},
    \label{eq:heat-expected-score-rate}
\end{equation}
uniformly in $T>0$. The constant $C$ depends only on $d,s,m,M,L$ and
the fixed comparison constants in the choice of $N_n$.
Here $|\Lambda_N|\asymp_d N^d$.
\end{theorem}

Since $|\Lambda_N|\asymp_d N^d$, the leading terms are $N^{-2s}$ from
approximation and $N^d/n$ from estimation. Thus Fourier OISM exhibits
the usual finite-dimensional approximation--estimation tradeoff.
Sections~\ref{subsec:heat-integrated-score-error} and
\ref{subsec:heat-variance-estimation} establish these two terms,
respectively. Section~\ref{subsec:heat-density-kl} balances them and
transfers the integrated score bound to minimax-optimal density
estimation under Kullback--Leibler loss.

\subsection{Integrated approximation error}
\label{subsec:heat-integrated-score-error}

We first prove the approximation term in
Theorem~\ref{thm:heat-expected-score-risk}. Assume throughout this
subsection that $p_0\in\mathcal P_s(m,M,L)$.

The gradient of $f_{t,N}^\star$ is the $L^2(\rho_t)$-orthogonal projection of $s_t$ onto
$\mathcal F_N$. We measure the approximation error by
\begin{equation}
    B_N^{\mathrm{heat}}(t)
        :=\norm{\nabla f_{t,N}^\star-s_t}_{L^2(\rho_t)}^2
        =\inf_{f\in\mathcal U_N}
            \norm{\nabla f-s_t}_{L^2(\rho_t)}^2.
    \label{eq:heat-population-approximation}
\end{equation}

\begin{proposition}[Integrated approximation error]
\label{prop:heat-integrated-approximation}
For every $p_0\in\mathcal P_s(m,M,L)$, $N\geq1$, and $T>0$,
\begin{equation}
    \int_0^T B_N^{\mathrm{heat}}(t)\,\mathrm{d}t\leq CN^{-2s},
    \label{eq:heat-integrated-approximation}
\end{equation}
where $C=C(d,s,m,M,L)$ is independent of $T$ and $N$.
\end{proposition}

The rate $N^{-2s}$ comes directly from the Fourier tail. Writing
$u_t:=\log p_t$ and letting $\widehat u_t(k)$ denote its Fourier
coefficients, Lemma~\ref{lem:heat-fourier-approximation} gives
\[
    \begin{aligned}
    B_N^{\mathrm{heat}}(t)
        &\leq M\sum_{|k|_2>N}|k|_2^2|\widehat u_t(k)|^2\\
        &\leq MN^{-2s}\sum_{i_0,\ldots,i_s=1}^d
            \norm{\partial_{i_0}\cdots\partial_{i_s}u_t}_{L^2(\nu)}^2.
    \end{aligned}
\]
The right-hand side involves $s+1$ derivatives of $\log p_t$, although
the initial density has only $s$ weak derivatives. Derivatives of $p_t$
through order $s$ are uniformly controlled over the chosen smoothness
class. The semigroup identity in Lemma~\ref{lem:heat-energy} controls the
remaining derivative after time integration. Lemma~\ref{lem:heat-integrated-approximation}
combines these facts to give a bound uniform in $T$, yielding
\eqref{eq:heat-integrated-approximation}. Appendix~\ref{subsec:heat-integrated-approximation}
gives the details.

\subsection{Variance and finite-sample estimation}
\label{subsec:heat-variance-estimation}

We now turn to statistical error. Unlike the Hermite products in
Section~\ref{sec:polynomial}, Fourier multiplication does not introduce
coefficients that grow with the cutoff. The resulting integrated variance
is controlled at the scale of the number of retained frequencies,
$|\Lambda_N|\asymp_d N^d$.
As in the OU setting, the estimation error is governed by fluctuations
of the empirical normal equations around their population counterpart.
Since Section~\ref{sec:heat} considers only the operator-informed estimator,
we suppress the superscript $\mathrm{OI}$ on its residual and covariance.
Define
\[
    \psi_{t,N}(x):=\mathfrak b_t(x)-\mathfrak a_t(x)c_{t,N}^\star,
    \qquad
    \Sigma_{t,N}:=\E_{\rho_0}\!\left[
        \psi_{t,N}(X_0)\psi_{t,N}(X_0)^\dagger\right].
\]
The population normal equations give
$\E_{\rho_0}\psi_{t,N}(X_0)=0$. The asymptotic
$L^2(\rho_t)$ variance is
\[
    V_N^{\mathrm{OI}}(t)
        =\tr\!\left(A_{t,N}^{-1}\Sigma_{t,N}\right).
\]

\begin{proposition}[Integrated variance and estimation bounds]
\label{prop:heat-variance-estimation}
Assume $m\leq p_0\leq M$. For every $N\geq1$, $n\geq1$, and $T>0$,
\begin{equation}
    \int_0^T V_N^{\mathrm{OI}}(t)\,\mathrm{d}t
        \leq\frac{|\Lambda_N|}{m}(1+2\log M).
    \label{eq:heat-integrated-variance-bound}
\end{equation}
Moreover,
\[
    \mathbb P_{p_0}(\mathcal C_N^c)
        \leq|\Lambda_N|\exp\!\left\{
            -\frac{m(1-\log2)n}{2|\Lambda_N|}
        \right\},
\]
and
\begin{equation}
    \E_{p_0}\!\left[1_{\mathcal C_N}
        \int_0^T
        \norm{\widehat s_{t,N}-\nabla f_{t,N}^\star}_{L^2(\rho_t)}^2
        \,\mathrm{d}t\right]
    \leq\frac{4M|\Lambda_N|}{m^2n}(1+2\log M).
    \label{eq:heat-expected-estimation-bound}
\end{equation}
Here $|\Lambda_N|\asymp_d N^d$.
\end{proposition}

The population comparison $A_{t,N}\succeq mK_N$ implies
$A_{t,N}^{-1}\preceq m^{-1}K_N^{-1}$. Hence the asymptotic variance
satisfies
\[
    V_N^{\mathrm{OI}}(t)
        =\tr\!\left(A_{t,N}^{-1}\Sigma_{t,N}\right)
        \leq\frac1m\tr\!\left(K_N^{-1}\Sigma_{t,N}\right).
\]
The same normalized residual covariance controls the finite-sample error:
\[
    \E_{p_0}\!\left[1_{\mathcal C_N}
        \norm{\widehat s_{t,N}-\nabla f_{t,N}^\star}_{L^2(\rho_t)}^2
        \right]
        \leq\frac{4M}{m^2n}
            \tr\!\left(K_N^{-1}\Sigma_{t,N}\right).
\]
Thus the asymptotic variance and $n$ times the finite-sample estimation
error are governed by the same normalized residual covariance. Its
integrated bound is
\[
    \int_0^T\tr(K_N^{-1}\Sigma_{t,N})\,\mathrm{d}t
        \leq|\Lambda_N|(1+2\log M).
\]
Integrating the preceding two inequalities therefore gives an integrated
asymptotic variance bound of order $|\Lambda_N|\asymp_d N^d$ and a finite-sample
estimation error bound of order $|\Lambda_N|/n\asymp_d N^d/n$. There is
therefore no analogue of the $4^N$ amplification from
Section~\ref{sec:polynomial}. The covariance estimate combines explicit
semigroup damping of the Fourier modes with entropy dissipation, while
matrix Chernoff and the Schur product theorem control the conditioning
event $\mathcal C_N$. Appendix~\ref{subsec:heat-estimation-proof} gives
the details.

On $\mathcal C_N$, orthogonality of the population projection gives the
exact decomposition
\[
    \norm{\widehat s_{t,N}-s_t}_{L^2(\rho_t)}^2
        =B_N^{\mathrm{heat}}(t)
          +\norm{\widehat s_{t,N}-\nabla f_{t,N}^\star}_{L^2(\rho_t)}^2.
\]
Propositions~\ref{prop:heat-integrated-approximation} and
\ref{prop:heat-variance-estimation} bound these two terms. On
$\mathcal C_N^c$, the estimator is zero and
$\mathcal R_T(0)\leq\log M$. Together with the conditioning-event bound,
these facts yield Theorem~\ref{thm:heat-expected-score-risk}; its proof
is given in Appendix~\ref{subsec:heat-estimation-proof}.

\subsection{Estimation rate and minimax optimality}
\label{subsec:heat-density-kl}

Theorem~\ref{thm:heat-expected-score-risk} gives leading integrated
score-estimation terms $N^{-2s}+N^d/n$. Choosing
$N_n\asymp n^{1/(2s+d)}$ balances them and yields
\[
    \sup_{p_0\in\mathcal P_s(m,M,L)}
        \E_{p_0}\mathcal R_T(\widehat s_{\cdot,N_n})
        \lesssim n^{-2s/(2s+d)}
\]
for all sufficiently large $n$, uniformly in $T>0$.

The integrated score loss also controls the distribution generated by
the approximate reverse diffusion. Conditional on the training data,
consider
\begin{equation}
    \mathrm{d}\widehat X_t^{\leftarrow}
        =2\widehat s_{T-t,N}(\widehat X_t^{\leftarrow})\,\mathrm{d}t
            +\sqrt2\,\mathrm{d}B_t,
    \qquad \widehat X_0^{\leftarrow}\sim\nu,
    \qquad 0\leq t\leq T,
    \label{eq:heat-learned-reverse}
\end{equation}
interpreted modulo $2\pi$, with the initial state and Brownian motion
independent of the training data. Let $\widehat p_{N,T}$ be the density
with respect to $\nu$ of the conditional law of
$\widehat X_T^{\leftarrow}$. For densities, write
$\mathrm{KL}(p\|q):=\int p\log(p/q)\,\mathrm{d}\nu$. Here $1$ denotes the
constant density of $\nu$. The Girsanov comparison gives
\begin{equation}
    \mathrm{KL}(p_0\|\widehat p_{N,T})
        \leq \mathrm{KL}(p_T\|1)
            +\mathcal R_T(\widehat s_{\cdot,N}).
    \label{eq:heat-kl-transfer}
\end{equation}
The first term is the initialization error from starting the learned
reverse process at $\nu$ instead of $\rho_T$. Heat-flow contraction gives
\[
    \mathrm{KL}(p_T\|1)\leq(M-1)e^{-2T}.
\]
Appendix~\ref{subsec:heat-kl-minimax-proof} proves the KL comparison
and this initialization bound.

\begin{corollary}[Minimax rate under Kullback--Leibler loss]
\label{cor:heat-density-kl}
For the fixed class~\eqref{eq:heat-density-class}, set
\[
    N_n:=\left\lfloor n^{1/(2s+d)}\right\rfloor,
    \qquad T_n:=\frac{s}{2s+d}\log n.
\]
Define $\widehat p_n:=\widehat p_{N_n,T_n}$ using the score estimator
$\widehat s_{\cdot,N_n}$ in~\eqref{eq:heat-learned-reverse}. There exists
a constant $C>0$, depending only on $d,s,m,M,L$, such that, for all
sufficiently large $n$,
\begin{equation}
    \sup_{p_0\in\mathcal P_s(m,M,L)}
        \E_{p_0}\mathrm{KL}(p_0\|\widehat p_n)
        \leq Cn^{-2s/(2s+d)}.
    \label{eq:heat-density-kl-upper}
\end{equation}
This rate is minimax optimal over $\mathcal P_s(m,M,L)$ among all
measurable density-valued estimators based on $n$ independent observations
from $\rho_0$.
\end{corollary}

Appendix~\ref{subsec:heat-kl-minimax-proof} proves the matching lower
bound by a localized Assouad construction on the torus.
The heat-flow setting therefore exhibits the standard
approximation--estimation tradeoff. Fourier truncation contributes
$N^{-2s}$, while estimation contributes $N^d/n$, and balancing the two
yields the classical nonparametric rate. This exponent agrees with the
corresponding squared total-variation rate for Euclidean Gaussian heat
flow in Dou et~al.\ \citep{dou2024optimal}, although our guarantee is
formulated under expected Kullback--Leibler loss.
The OISM coefficients are
computed directly from the original observations, so the minimax
density-estimation rate is attained without simulating the forward
diffusion.

\section{Discussion and open questions}
\label{sec:discussion}

The comparison developed above suggests that the statistical behavior of
score matching over growing approximation families depends not only on
approximation quality, but also on algebraic interactions among the chosen
features. In the two examples studied here, this distinction appears
through the product structure of generator eigenfunctions. It would be
useful to identify general conditions on a diffusion generator and its
eigenfunctions under which increasing cutoffs lead to controlled rather
than exponential growth of estimation error. Such a theory could guide
the choice of diffusion-adapted approximation families for score estimation.

Our OU results leave an important statistical question open. The variance
lower bounds first take the sample size to infinity at fixed cutoff $N$,
and then study the resulting asymptotic variance as $N\to\infty$. They
therefore do not imply a finite-sample lower bound when the cutoff
$N=N(n)$ grows with the sample size. Establishing such a result would make
precise the sample complexity consequences of the Hermite-product
amplification. It would also be interesting to determine whether a
different choice of approximation space can avoid this instability while
retaining comparable approximation guarantees for the score.

The heat-flow analysis also suggests several extensions. It is natural
to ask whether analogous minimax guarantees hold for other compact state
spaces, generators, or approximation schemes with comparably tractable
semigroup and product structure. Our analysis treats the learned reverse
diffusion in continuous time and does not account for numerical
discretization of the reverse SDE. An end-to-end guarantee combining
score-estimation and discretization error would therefore be a natural
next step. More generally, identifying which combinations of diffusion,
approximation family, and score-matching objective yield controlled
estimation error as the approximation space grows may provide a useful
principle for their design.

\medskip
\noindent\textbf{Acknowledgement.} The author thanks Sinho Chewi and Aram-Alexandre 
Pooladian for generously devoting their time to many helpful discussions.

\bibliographystyle{alphaurl}
\bibliography{references}

\newpage
\appendix
\numberwithin{equation}{section}

\section{Supporting derivations}
\label{app:supporting-derivations}

\subsection{Score matching and asymptotic variance}
\label{subsec:score-matching-asymptotics}

Integration by parts gives
\[
    \E_{\rho_t}\langle\nabla f,s_t\rangle
        =-\E_{\rho_t}\Delta f.
\]
Expanding out the square in the score-matching objective therefore gives the identity in
Section~\ref{subsec:score-matching}. For the finite-dimensional calculation,
choose a real basis $\{u_j:j\in I\}$ of $\mathcal U$ modulo constants
and write $f_c=\sum_{j\in I}c_j u_j$. Set
\[
    \mathfrak a(x)_{jk}:=\langle\nabla u_j(x),\nabla u_k(x)\rangle,
    \qquad \mathfrak b(x)_j:=-\Delta u_j(x).
\]
Then $\ell_{f_c}(x)=\frac12 c^\top\mathfrak a(x)c-
\mathfrak b(x)^\top c$. Write $A_t:=\E_{\rho_t}\mathfrak a(X_t)$,
$b_t:=\E_{\rho_t}\mathfrak b(X_t)$, and
\[
    c_t^\star:=A_t^{-1}b_t,\qquad
    \psi_t(x):=\mathfrak b(x)-\mathfrak a(x)c_t^\star,\qquad
    \Sigma_t:=\E_{\rho_t}[\psi_t(X_t)\psi_t(X_t)^\top].
\]
Let $\widehat A_t$ and $\widehat b_t$ be the corresponding sample
averages, $\widehat c_t:=\widehat A_t^{-1}\widehat b_t$, and
$f_t^\star:=f_{c_t^\star}$, $\widehat f_t:=f_{\widehat c_t}$,
$\widehat s_t:=\nabla\widehat f_t$.

For the asymptotic distribution, the population and empirical
minimizers satisfy
\[
    \widehat c_t-c_t^\star
        =\widehat A_t^{-1}\frac1n
            \sum_{i=1}^n\psi_t(X_t^{(i)}).
\]
Integrability of the entries of $\mathfrak a(X_t)$ gives
$\widehat A_t\to A_t$ in probability. Since $A_t$ is positive
definite, $\widehat A_t$ is invertible with probability tending to one
and $\widehat A_t^{-1}\to A_t^{-1}$ in probability. Assuming additionally
that $\E_{\rho_t}|\psi_t(X_t)|^2<\infty$, the multivariate central limit
theorem and Slutsky's theorem yield
\begin{equation}
    \sqrt n(\widehat c_t-c_t^\star)
    =A_t^{-1}\frac1{\sqrt n}
        \sum_{i=1}^n\psi_t(X_t^{(i)})+o_{\mathbb P}(1),
    \label{eq:appendix-linear-expansion}
\end{equation}
which gives the asymptotic normality of $\widehat c_t$. Moreover,
\begin{equation}
    \norm{\widehat s_t-\nabla f_t^\star}_{L^2(\rho_t)}^2
    =(\widehat c_t-c_t^\star)^\top A_t
        (\widehat c_t-c_t^\star)
    =2\{L_t(\widehat f_t)-L_t(f_t^\star)\}.
    \label{eq:appendix-score-excess-risk}
\end{equation}
Let $Z_t\sim\mathcal N(0,A_t^{-1}\Sigma_tA_t^{-1})$ denote the limiting
Gaussian vector. The continuous mapping theorem gives the limiting quadratic form
$Z_t^\top A_tZ_t$. When $\mathcal U=\mathcal U_N$, its mean is the
asymptotic variance $V_N(t)=\tr(A_t^{-1}\Sigma_t)$.

The same argument applies to OISM with the single-observation quantities
$P_t\mathfrak a$ and $P_t\mathfrak b$ and sampling law $\rho_0$.
Here $\psi_t^{\mathrm{OI}}:=P_t\psi_t$ and
$\Sigma_t^{\mathrm{OI}}:=\E_{\rho_0}[\psi_t^{\mathrm{OI}}(X_0)
\psi_t^{\mathrm{OI}}(X_0)^\top]$.
Conditional Jensen's inequality gives
\[
    \E_{\rho_0}\norm{\psi_t^{\mathrm{OI}}(X_0)}^2
        \leq \E_{\rho_t}\norm{\psi_t(X_t)}^2,
\]
so its limiting coefficient covariance is
$A_t^{-1}\Sigma_t^{\mathrm{OI}}A_t^{-1}$. The conditional covariance
identity then gives $V_N^{\mathrm{OI}}(t)\leq V_N(t)$.

\subsection{Complex Fourier coordinates}
\label{subsec:heat-complex-proof}

Let
\[
    \Lambda_N:=\{k\in\mathbb Z^d:0<|k|_2\leq N\},
    \qquad \omega_k(x):=e^{i\langle k,x\rangle}.
\]
A real trigonometric potential can be written as
$f_c=\sum_{k\in\Lambda_N}c_k\omega_k$ with
$c_{-k}=\overline{c_k}$. For one representative of each pair
$\{k,-k\}$, writing $c_k=a_k+ib_k$ gives
\[
    c_k\omega_k+c_{-k}\omega_{-k}
        =2a_k\cos\langle k,x\rangle
            -2b_k\sin\langle k,x\rangle,
\]
so these complex coordinates describe the same real potential space as
the sine--cosine parametrization.

For arbitrary $c\in\mathbb C^{\Lambda_N}$, define
$|\nabla f_c|^2:=\overline{\nabla f_c}\cdot\nabla f_c$. Since
\[
    \nabla f_c=\sum_{k\in\Lambda_N}ikc_k\omega_k,
    \qquad
    \Delta f_c=-\sum_{k\in\Lambda_N}|k|_2^2c_k\omega_k,
\]
the Fourier product rule and $P_t\omega_k=e^{-t|k|_2^2}\omega_k$ give
\begin{align*}
    P_t|\nabla f_c|^2(x)
        &=\sum_{k,\ell\in\Lambda_N}
            \overline{c_k}c_\ell(k\cdot\ell)
            e^{-t|\ell-k|_2^2}\omega_{\ell-k}(x),\\
    -P_t\Delta f_c(x)
        &=\sum_{k\in\Lambda_N}|k|_2^2e^{-t|k|_2^2}
            c_k\omega_k(x).
\end{align*}
Thus the operator-informed loss is a Hermitian quadratic form in $c$.
When $c$ is conjugate symmetric, it agrees with the real-coordinate loss.
Conversely, if $f_c=u+iv$ with $u,v$ real, then
\[
    \norm{\nabla f_c-s_t}_{L^2(\rho_t)}^2
        =\norm{\nabla u-s_t}_{L^2(\rho_t)}^2
            +\norm{\nabla v}_{L^2(\rho_t)}^2.
\]
Complexification therefore cannot improve the population minimum.
The population and empirical quadratic objectives are invariant under the
involution $c^\sharp_k:=\overline{c_{-k}}$. Uniqueness of a minimizer forces
$c=c^\sharp$, hence conjugate symmetry, so the fitted potential is real-valued.
Identifying conjugate-symmetric coefficient vectors with the corresponding
real sine--cosine coordinates, the asymptotic calculation of
Appendix~\ref{subsec:score-matching-asymptotics} applies unchanged.
Equivalently, in the complex representation, the same formulas hold with
transposes replaced by conjugate transposes, yielding the trace expression
used in Section~\ref{sec:heat} without an additional factor of two.

\section{Proofs for polynomial approximation under OU flow}
\label{app:polynomial-proofs}

This appendix supplies the technical details for the variance-growth results
summarized in Section~\ref{subsec:ou-instability-mechanism}. We first record
the coefficient and approximation bounds used there, then prove the
high-degree residual lower bound and the analytic obstruction to exponential
approximation decay. We conclude with the proofs of
Theorems~\ref{thm:fixed-law-variance-growth}
and~\ref{thm:oi-polynomial-variance-growth}.
It also proves the local minimax lower bound from
Section~\ref{subsec:ou-local-minimax}.

For fixed $N$, abbreviate
$\psi_t:=\psi_{t,N}$, $\Sigma_t:=\Sigma_{t,N}$, and
$\psi_t^{\mathrm{OI}}:=\psi_{t,N}^{\mathrm{OI}}$.
Throughout this section, $\rho_0$ satisfies
\eqref{eq:polynomial-density-ratio}. The Markov property gives
$m\leq r_t=P_tr_0\leq M$. For $t>0$, Gaussian smoothing gives
(with $\mathrm{id}(x)=x$)
\begin{equation}
    \norm{s_t+\mathrm{id}}_\infty
    =\norm{\frac{r_t'}{r_t}}_\infty
    \leq\frac{M}{m}\frac{e^{-t}}{\sqrt{1-e^{-2t}}}\E|Z|,
    \qquad Z\sim\mathcal N(0,1).
    \label{eq:score-smoothing}
\end{equation}
Indeed, Mehler's formula, followed by differentiation of its Gaussian
kernel, gives
\[
    r_t(x)=\E\!\left[
        r_0\!\left(e^{-t}x+\sqrt{1-e^{-2t}}Z\right)\right],
    \qquad
    r_t'(x)=\frac{e^{-t}}{\sqrt{1-e^{-2t}}}
        \E\!\left[
            Zr_0\!\left(e^{-t}x+\sqrt{1-e^{-2t}}Z\right)\right].
\]
Thus $r_0\leq M$ bounds $\norm{r_t'}_\infty$ by the right-hand side
of~\eqref{eq:score-smoothing} without the factor $1/m$, and division by
$r_t\geq m$ proves the estimate. This also justifies the required score
integrability and integration by parts at positive times.

\subsection{Population coefficients and approximation bounds}

We first bound the highest fitted Hermite coefficient and compare the
$L^2(\rho_t)$ and $L^2(\pi)$ approximation errors.

Set $f_{t,N}^\star:=f_{c_{t,N}^\star}$ and define
\[
    D_N:=\operatorname{diag}(1,\ldots,N),
    \qquad G_t:=D_N^{-1/2}A_{t,N}D_N^{-1/2}.
\]
The normalization by $D_N$ accounts for the factors $\sqrt j$ in
$\phi_j'=\sqrt j\,\phi_{j-1}$. If $\rho_0=\pi$, then $\rho_t=\pi$, and
Hermite orthonormality gives $A_{t,N}=D_N$ and thus $G_t=I$. More generally,
for $z\in\mathbb R^N$, Hermite orthogonality and
$m\leq r_t\leq M$ give
\[
    m\norm{z}_2^2\leq z^\top G_tz\leq M\norm{z}_2^2.
\]
Let
\[
    \mu_k(t):=\E_{\rho_t}\phi_k(X_t)
        =\langle r_t,\phi_k\rangle_{L^2(\pi)}.
\] We have
\[
    (b_{t,N})_1=0,\qquad
    (b_{t,N})_j=-\sqrt{j(j-1)}\,\mu_{j-2}(t),
    \quad j\geq2.
\]
Consequently,
\[
    \begin{aligned}
    \norm{D_N^{-1/2}b_{t,N}}_2^2
        &=\sum_{j=2}^N(j-1)\mu_{j-2}(t)^2\\
        &\leq N\sum_{k\geq0}\mu_k(t)^2
        =N\norm{r_t}_{L^2(\pi)}^2
        \leq MN.
    \end{aligned}
\]
Here the last inequality follows from $0\leq r_t\leq M$ and
$\int r_t\,\mathrm{d}\pi=1$. Since $A_{t,N}c_{t,N}^\star=b_{t,N}$,
\[
    D_N^{1/2}c_{t,N}^\star
        =G_t^{-1}D_N^{-1/2}b_{t,N}.
\]
It follows that
\begin{equation}
    N\alpha_N(t)^2
        \leq\norm{D_N^{1/2}c_{t,N}^\star}_2^2
        \leq\frac{MN}{m^2},
    \qquad t>0.
    \label{eq:score-top-coefficient-upper}
\end{equation}
If $s_t=\sum_{k\geq0}\gamma_k(t)\phi_k$ in $L^2(\pi)$, also set
\begin{equation}
    H_N(t):=\inf_{w\in\mathcal P_{N-1}}
        \norm{s_t-w}_{L^2(\pi)}^2
        =\sum_{k\geq N}|\gamma_k(t)|^2.
    \label{eq:gaussian-score-tail}
\end{equation}
Since the Hermite polynomials are dense in $L^2(\pi)$,
$H_N(t)\downarrow0$. Equivalence of the $L^2(\pi)$ and
$L^2(\rho_t)$ norms gives
\[
    mH_N(t)\leq B_N^{\mathrm{OU}}(t)\leq MH_N(t),
\]
and therefore $B_N^{\mathrm{OU}}(t)\downarrow0$.

\subsection{High-degree residual amplification}

The normalized Hermite product identity used in
Section~\ref{sec:oi-polynomial} is
\[
    \phi_j\phi_k
        =\sum_{r=0}^{\min(j,k)}
            r!\binom{j}{r}\binom{k}{r}
            \sqrt{\frac{(j+k-2r)!}{j!k!}}\,\phi_{j+k-2r}.
\]

\begin{lemma}[Highest-degree contribution]
\label{lem:polynomial-amplification}
For every $N\geq3$ and $t>0$,
\begin{equation}
    \begin{aligned}
    V_N(t)
        &\geq\frac{m}{M}N\alpha_N(t)^2
            \binom{2N-2}{N-1}\\
        &\geq\frac{m}{M^2}\Delta_N(t)
            \binom{2N-2}{N-1}.
    \end{aligned}
    \label{eq:polynomial-coefficient-bound}
\end{equation}
\end{lemma}

\begin{proof}
The $N$th coordinate of the score residual is
\[
    (\psi_t(x))_N
        =-\sqrt{N(N-1)}\,\phi_{N-2}(x)
        -\sum_{k=1}^N\sqrt{Nk}\,(c_{t,N}^\star)_k
            \phi_{N-1}(x)\phi_{k-1}(x).
\]
The leading coefficient of $\phi_j$ is $1/\sqrt{j!}$, so
\begin{equation}
    \phi_{N-1}^2
        =\sqrt{\binom{2N-2}{N-1}}\,\phi_{2N-2}
            +\text{terms of degree at most }2N-4.
    \label{eq:hermite-leading-product}
\end{equation}
For $k<N$, the product in the sum has degree at most $2N-3$, while
the linear term has degree $N-2$. Thus the coefficient of
$\phi_{2N-2}$ in $(\psi_t)_N$ is
\[
    -N\alpha_N(t)\sqrt{\binom{2N-2}{N-1}}.
\]
Hermite orthogonality, $r_t\geq m$, and centering of the residual give
\[
    (\Sigma_t)_{NN}
        \geq mN^2\alpha_N(t)^2\binom{2N-2}{N-1}.
\]
Define
\[
    \langle u,v\rangle_{A_{t,N}}:=u^\top A_{t,N}v.
\]
Since $A_{t,N}$ is positive definite, this is an inner product. For any vector
$z$, apply Cauchy--Schwarz in this inner product to $A_{t,N}^{-1}z$ and the
$N$th standard basis vector $e_N$. We have
\[
    \langle A_{t,N}^{-1}z,e_N\rangle_{A_{t,N}}
        =(A_{t,N}^{-1}z)^\top A_{t,N}e_N=z_N,
\]
and hence
\[
    z_N^2
        \leq
        \langle A_{t,N}^{-1}z,A_{t,N}^{-1}z\rangle_{A_{t,N}}
        \langle e_N,e_N\rangle_{A_{t,N}}
        =\bigl(z^\top A_{t,N}^{-1}z\bigr)(A_{t,N})_{NN}.
\]
Therefore $z^\top A_{t,N}^{-1}z\geq z_N^2/(A_{t,N})_{NN}$. Taking
$z=\psi_t(X_t)$ and expectations gives
\[
    V_N(t)
        =\E_{\rho_t}\!\left[
            \psi_t(X_t)^\top A_{t,N}^{-1}\psi_t(X_t)\right]
        \geq
        \frac{\E_{\rho_t}\!\left[(\psi_t(X_t))_N^2\right]}
            {(A_{t,N})_{NN}}.
\]
Because the residual is centered,
$\E_{\rho_t}\psi_t(X_t)=0$, so
\[
    \E_{\rho_t}\!\left[(\psi_t(X_t))_N^2\right]
        =(\Sigma_t)_{NN}.
\]
Thus $V_N(t)\geq(\Sigma_t)_{NN}/(A_{t,N})_{NN}$. Finally,
\[
    (A_{t,N})_{NN}=N\E_{\rho_t}\phi_{N-1}^2\leq MN.
\]
Combining this estimate with the preceding lower bound on
$(\Sigma_t)_{NN}$ proves the first inequality.
The second follows from Lemma~\ref{lem:bias-increments}.
\end{proof}

\begin{proof}[Proof of Proposition~\ref{prop:ou-residual-amplification}]
The direct bound follows from Lemma~\ref{lem:polynomial-amplification}.
The $N$th coordinate of the operator-informed residual is
\[
    (\psi_t^{\mathrm{OI}}(x))_N
        =-\sqrt{N(N-1)}e^{-(N-2)t}\phi_{N-2}(x)
        -\sum_{k=1}^N\sqrt{Nk}\,(c_{t,N}^\star)_k
            P_t(\phi_{N-1}\phi_{k-1})(x).
\]
For $N\geq3$, the first term has degree $N-2$, and the product term
has degree at most $2N-3$ when $k<N$. By
\eqref{eq:hermite-leading-product} and
$P_t\phi_j=e^{-jt}\phi_j$, the coefficient at $\phi_{2N-2}$ is
\[
    -N\alpha_N(t)e^{-2(N-1)t}
        \sqrt{\binom{2N-2}{N-1}}.
\]
Semigroup averaging multiplies the degree-$(2N-2)$ Hermite component
of the direct residual by $e^{-(2N-2)t}$, so its squared contribution
gains the factor $e^{-4(N-1)t}$. The residual is centered under $\rho_0$,
where $r_0\geq m$, and $(A_{t,N})_{NN}\leq MN$. The same Cauchy--Schwarz
comparison in the $A_{t,N}$-inner product as in
Lemma~\ref{lem:polynomial-amplification}, together with
Lemma~\ref{lem:bias-increments}, gives
\begin{equation}
    \begin{aligned}
    V_N^{\mathrm{OI}}(t)
        &\geq\frac mM N\alpha_N(t)^2e^{-4(N-1)t}
            \binom{2N-2}{N-1}\\
        &\geq\frac{m}{M^2}\Delta_N(t)e^{-4(N-1)t}
            \binom{2N-2}{N-1}.
    \end{aligned}
    \label{eq:oi-coefficient-bound}
\end{equation}
\end{proof}

\subsection{Analytic obstruction}

The following lemmas supply the analytic obstruction used in
Lemma~\ref{lem:non-exponential-increments}.

\begin{lemma}[Entire extension from coefficient decay]
\label{lem:hermite-entire}
Suppose
\[
    s=\sum_{k\geq0}\gamma_k\phi_k
    \qquad\text{in }L^2(\pi),
\]
and $|\gamma_k|\leq Ce^{-\eta k}$ for some $C,\eta>0$. Then $s$ agrees
$\pi$-almost everywhere on $\mathbb R$ with the restriction of an entire
function on $\mathbb C$.
\end{lemma}

\begin{proof}
The generating function is
\[
    e^{zw-w^2/2}=\sum_{k\geq0}\frac{\phi_k(z)}{\sqrt{k!}}w^k.
\]
Fix $R>0$. For $|z|\leq R$ and $r>0$, Cauchy's coefficient estimate gives
\[
    \frac{|\phi_k(z)|}{\sqrt{k!}}
        \leq r^{-k}\max_{|w|=r}|e^{zw-w^2/2}|
        \leq r^{-k}e^{Rr+r^2/2}.
\]
Taking $r=\sqrt k$ for $k\geq1$ yields
\[
    |\phi_k(z)|
        \leq\sqrt{k!}\,k^{-k/2}e^{R\sqrt k+k/2}.
\]
Stirling's formula therefore gives, uniformly for $|z|\leq R$,
\[
    |\phi_k(z)|\leq C_Rk^{1/4}e^{R\sqrt k}.
\]
Consequently,
\[
    |\gamma_k\phi_k(z)|
        \leq C\cdot C_Rk^{1/4}e^{-\eta k+R\sqrt k}.
\]
For all sufficiently large $k$,
$-\eta k+R\sqrt k\leq-\eta k/2$, so the right-hand side is summable in
$k$. The Weierstrass $M$-test gives local uniform convergence on
$\mathbb C$, and
\[
    F(z):=\sum_{k\geq0}\gamma_k\phi_k(z)
\]
is entire. Its partial sums converge to $s$ in $L^2(\pi)$, so
some subsequence converges to $s$ $\pi$-almost everywhere. The same
subsequence converges pointwise on $\mathbb R$ to $F$ by the local uniform
convergence above. Therefore, $F=s$ $\pi$-almost everywhere on $\mathbb R$.
\end{proof}

\begin{lemma}[The score is not entire]
\label{lem:analytic-obstruction}
If $r_0\not\equiv1$, then $s_t$ does not admit an entire extension for
any $t>0$.
\end{lemma}

\begin{proof}
Fix $t>0$. Mehler's formula extends $r_t$ to $\mathbb C$ by
\[
    r_t(z)=\frac1{\sqrt{2\pi(1-e^{-2t})}}
        \int_{\mathbb R}r_0(y)
        \exp\!\left(-\frac{(y-e^{-t}z)^2}{2(1-e^{-2t})}\right)\,\mathrm{d}y.
\]
The bound $r_0\leq M$ provides an integrable dominating function for the
integrand and its $z$-derivatives, uniformly on compact subsets of
$\mathbb C$. Differentiation under the integral is therefore justified,
so this extension is entire. Write $z=u+iv$. Then
\[
    y-e^{-t}z=(y-e^{-t}u)-ie^{-t}v,
\]
and
\[
    \operatorname{Re}\!\left(
        -\frac{(y-e^{-t}z)^2}{2(1-e^{-2t})}
    \right)
    =-\frac{(y-e^{-t}u)^2}{2(1-e^{-2t})}
        +\frac{e^{-2t}v^2}{2(1-e^{-2t})}.
\]
It follows that
\[
    \left|
        \exp\!\left(
            -\frac{(y-e^{-t}z)^2}{2(1-e^{-2t})}
        \right)
    \right|
    =
    \exp\!\left(
        -\frac{(y-e^{-t}u)^2}{2(1-e^{-2t})}
        +\frac{e^{-2t}v^2}{2(1-e^{-2t})}
    \right).
\]
Using $r_0\leq M$ in the Mehler representation therefore gives
\[
    \begin{aligned}
    |r_t(z)|
        \leq{}&M
        \exp\!\left(
            \frac{e^{-2t}v^2}{2(1-e^{-2t})}
        \right)
        \frac1{\sqrt{2\pi(1-e^{-2t})}}\\
        &\quad\cdot
        \int_{\mathbb R}
        \exp\!\left(
            -\frac{(y-e^{-t}u)^2}{2(1-e^{-2t})}
        \right)\,\mathrm{d}y.
    \end{aligned}
\]
The integral, together with its Gaussian normalization factor, equals one, and therefore
\[
    |r_t(z)|
        \leq M
        \exp\!\left(
            \frac{e^{-2t}v^2}{2(1-e^{-2t})}
        \right).
\]
Since $v^2\leq|z|^2$,
\[
    |r_t(z)|
        \leq M
        \exp\!\left(
            \frac{e^{-2t}|z|^2}{2(1-e^{-2t})}
        \right).
\]
Thus $r_t$ has order at most two.

Suppose first that $r_t$ has no complex zero. The zero-free case of
Hadamard factorization \citep[Ch.~XI, \S3]{conway1978functions} gives
$r_t=e^Q$ for a polynomial $Q$ of degree at most two. For $x\in\mathbb R$,
\[
    r_t(x)=e^{Q(x)}
        =e^{\operatorname{Re}Q(x)}e^{i\operatorname{Im}Q(x)}.
\]
Since $r_t(x)>0$, we have $e^{i\operatorname{Im}Q(x)}=1$, and hence
\[
    \operatorname{Im}Q(x)\in2\pi\mathbb Z,
    \qquad x\in\mathbb R.
\]
The function $x\mapsto\operatorname{Im}Q(x)$ is continuous, so the
connectedness of $\mathbb R$ implies that it is constant. Moreover,
$m\leq r_t(x)\leq M$ on the real axis and
$r_t(x)=e^{\operatorname{Re}Q(x)}$, so
\[
    \log m\leq\operatorname{Re}Q(x)\leq\log M,
    \qquad x\in\mathbb R.
\]
Writing $Q(z)=az^2+bz+c$, the constancy of
$\operatorname{Im}Q(x)$ on $\mathbb R$ implies that $a$ and $b$ are real,
while the boundedness of
$\operatorname{Re}Q(x)=ax^2+bx+\operatorname{Re}(c)$ on $\mathbb R$
forces $a=b=0$. Thus $Q$ is constant, and normalization gives
$r_t\equiv1$. Since
$P_t\phi_k=e^{-kt}\phi_k$ and $r_0\in L^2(\pi)$, injectivity of $P_t$
then gives $r_0\equiv1$, a contradiction. Therefore $r_t$ has a
complex zero.

Suppose that an entire function $S$ agrees with $s_t$ almost everywhere
on $\mathbb R$. Since both functions are continuous there, they agree
everywhere on $\mathbb R$. The identity $p_t=r_tp_\pi$ gives
\[
    \frac{r_t'(x)}{r_t(x)}=s_t(x)+x=S(x)+x,
    \qquad x\in\mathbb R.
\]
The entire function
\[
    F(z):=r_t'(z)-(S(z)+z)r_t(z)
\]
therefore vanishes on $\mathbb R$. The identity theorem gives
\[
    r_t'(z)=(S(z)+z)r_t(z),
    \qquad z\in\mathbb C.
\]
Since $S(z)+z$ has an entire primitive,
\[
    r_t(z)=r_t(0)
        \exp\!\left(\int_0^z(S(w)+w)\,\mathrm{d}w\right),
\]
which is zero-free. This contradiction proves the lemma.
\end{proof}

\subsection{Remaining variance-growth proofs}

We now combine the preceding technical ingredients to prove the fixed-time
variance bounds.

\begin{proof}[Proof of Lemma~\ref{lem:non-exponential-increments}]
Fix $t>0$. Since $0\leq\Delta_N(t)\leq B_{N-1}^{\mathrm{OU}}(t)$,
the approximation increments are uniformly bounded, so the limit superior
in~\eqref{eq:non-exponential-increments} is at most zero. If it were
strictly less than zero, then $\Delta_N(t)\leq Ce^{-\eta N}$ for some
$C,\eta>0$ and all sufficiently large $N$. Since
$B_N^{\mathrm{OU}}(t)\downarrow0$, telescoping would give
\[
    B_N^{\mathrm{OU}}(t)=\sum_{j>N}\Delta_j(t)
        \leq C'e^{-\eta N}.
\]
By~\eqref{eq:gaussian-score-tail},
$|\gamma_N(t)|^2\leq H_N(t)\leq B_N^{\mathrm{OU}}(t)/m$.
Thus the Hermite coefficients of $s_t$ decay exponentially, and
Lemma~\ref{lem:hermite-entire} would give an entire extension of $s_t$.
This contradicts Lemma~\ref{lem:analytic-obstruction}, proving
\eqref{eq:non-exponential-increments}.
\end{proof}

\begin{proof}[Proof of Theorem~\ref{thm:fixed-law-variance-growth}]
Fix $t>0$. Proposition~\ref{prop:ou-residual-amplification} gives
\[
    V_N(t)\gtrsim_{m,M}\Delta_N(t)\binom{2N-2}{N-1}.
\]
By Stirling's formula, $\binom{2N-2}{N-1}=4^{N-o(N)}$. Taking
$N^{-1}\log$ and the limit superior, Lemma~\ref{lem:non-exponential-increments}
therefore yields
\[
    \limsup_{N\to\infty}\frac1N\log V_N(t)\geq\log4,
\]
proving~\eqref{eq:fixed-law-variance-growth}.
\end{proof}

\begin{proof}[Proof of Theorem~\ref{thm:oi-polynomial-variance-growth}]
Fix $t>0$. Proposition~\ref{prop:ou-residual-amplification} gives
\[
    V_N^{\mathrm{OI}}(t)\gtrsim_{m,M}
        \Delta_N(t)e^{-4(N-1)t}\binom{2N-2}{N-1}.
\]
Taking $N^{-1}\log$ and the limit superior, using
$\binom{2N-2}{N-1}=4^{N-o(N)}$ and
Lemma~\ref{lem:non-exponential-increments}, gives
\[
    \limsup_{N\to\infty}\frac1N\log V_N^{\mathrm{OI}}(t)
        \geq\log4-4t,
\]
proving~\eqref{eq:oi-pointwise-variance-growth}.
\end{proof}

\subsection{Local minimax lower bound}
\label{subsec:ou-local-minimax-proof}

\begin{proof}[Proof of Theorem~\ref{thm:ou-local-minimax}]
Fix $t>0$ and $N\geq1$. Write
$c^\star:=c_{t,N}^\star$, $\psi^{\mathrm{OI}}:=\psi_{t,N}^{\mathrm{OI}}$,
$\Sigma^{\mathrm{OI}}:=\Sigma_{t,N}^{\mathrm{OI}}$, and
$\Gamma:=A_{t,N}^{-1/2}\Sigma^{\mathrm{OI}}A_{t,N}^{-1/2}$. If $\Gamma=0$, then
$V_N^{\mathrm{OI}}(t)=0$ and the result is immediate. Otherwise, let
$q:=\operatorname{rank}(\Gamma)$ and let $v_1,\ldots,v_q$ be orthonormal
eigenvectors of $\Gamma$ with positive eigenvalues
$\lambda_1,\ldots,\lambda_q$. Since the remaining eigenvalues vanish,
\[
    \sum_{j=1}^q\lambda_j
    =\tr(\Gamma)
    =\tr(A_{t,N}^{-1}\Sigma^{\mathrm{OI}})
    =V_N^{\mathrm{OI}}(t).
\]

For $X_0\sim\rho_0$, set
$Z_j:=v_j^\top A_{t,N}^{-1/2}\psi^{\mathrm{OI}}(X_0)$ and
$h_j:=Z_j/\sqrt{\lambda_j}$. The population normal equations give
$\E_{\rho_0}Z_j=0$, while
$\E_{\rho_0}[Z_jZ_k]=\lambda_j\delta_{jk}$. Thus the $h_j$ form an
orthonormal family in
$L_0^2(\rho_0):=\{h\in L^2(\rho_0):\E_{\rho_0}h=0\}$.

Bounded centered functions are dense in $L_0^2(\rho_0)$, so choose
$\widetilde h_1,\ldots,\widetilde h_q$ arbitrarily close to
$h_1,\ldots,h_q$ in $L^2(\rho_0)$. Define
$C_{k\ell}:=\E_{\rho_0}[Z_k\widetilde h_\ell]$. Since
$\widetilde h_\ell\to h_\ell$ in $L^2(\rho_0)$,
$C_{k\ell}\to\sqrt{\lambda_k}\,\delta_{k\ell}$, so $C$ is invertible
once the approximations are sufficiently close. Define the bounded centered
functions
\[
    g_j:=\sqrt{\lambda_j}\sum_{\ell=1}^q
        \widetilde h_\ell(C^{-1})_{\ell j},
    \qquad j=1,\ldots,q.
\]
Then
\[
    \E_{\rho_0}[Z_kg_j]
        =\sqrt{\lambda_j}\sum_{\ell=1}^q
            C_{k\ell}(C^{-1})_{\ell j}
        =\sqrt{\lambda_j}\,\delta_{kj}.
\]
To verify the norm bound, define
$B_{k\ell}:=\langle h_k,\widetilde h_\ell\rangle_{L^2(\rho_0)}$.
Since $Z_k=\sqrt{\lambda_k}\,h_k$, we have
$C_{k\ell}=\sqrt{\lambda_k}\,B_{k\ell}$ and therefore
$\sqrt{\lambda_j}(C^{-1})_{\ell j}=(B^{-1})_{\ell j}$. Hence
$g_j=\sum_{\ell=1}^q\widetilde h_\ell(B^{-1})_{\ell j}$.
As $B\to I_q$, it follows that $g_j\to h_j$ in $L^2(\rho_0)$.
Taking the approximations sufficiently close therefore ensures
$\E_{\rho_0}[g_j^2]\leq2$ for every $j$.

For $\theta\in\{-1,1\}^q$, set
$G_\theta:=\sum_{j=1}^q\theta_jg_j$ and
$K:=\max_\theta\norm{G_\theta}_\infty<\infty$. Fix a sufficiently small
numerical constant $\delta>0$, set $\varepsilon_n:=\delta/\sqrt n$, and
define
$\mathrm{d}\rho_{0,\theta,n}/\mathrm{d}\rho_0
:=1+\varepsilon_nG_\theta$. Centeredness gives total mass one, and
$\varepsilon_nK\to0$ gives positivity for all sufficiently large $n$.
Moreover,
$r_{0,\theta,n}=r_0(1+\varepsilon_nG_\theta)$. Since
$m+\eta\leq r_0\leq M-\eta$, the bounds
$m\leq r_{0,\theta,n}\leq M$ hold uniformly in $\theta$ for all sufficiently
large $n$. Also,
$\norm{r_{0,\theta,n}-r_0}_\infty
\leq(M-\eta)\varepsilon_nK\to0$, so $r_{0,\theta,n}\not\equiv1$ for all
sufficiently large $n$. Hence the $\rho_{0,\theta,n}$ satisfy
\eqref{eq:polynomial-density-ratio}, and
\[
    \max_\theta
    \left\|
        \frac{\mathrm{d}\rho_{0,\theta,n}}{\mathrm{d}\rho_0}-1
    \right\|_\infty
    \leq\frac{\delta K}{\sqrt n}.
\]

We next determine how the perturbations change the population polynomial
projection. For any bounded centered $g$, let
$\mathrm{d}\rho_{0,\varepsilon g}/\mathrm{d}\rho_0:=1+\varepsilon g$,
$A_\varepsilon:=\E_{\rho_{0,\varepsilon g}}[\mathfrak a_t(X_0)]$,
$b_\varepsilon:=\E_{\rho_{0,\varepsilon g}}[\mathfrak b_t(X_0)]$, and
$c_\varepsilon^\star:=A_\varepsilon^{-1}b_\varepsilon$. Differentiating
$A_\varepsilon c_\varepsilon^\star=b_\varepsilon$ at $\varepsilon=0$ and
using
$\psi^{\mathrm{OI}}=\mathfrak b_t-\mathfrak a_tc^\star$ gives
\[
    \left.\frac{\mathrm{d}}{\mathrm{d}\varepsilon}
        c_\varepsilon^\star\right|_{\varepsilon=0}
    =A_{t,N}^{-1}\E_{\rho_0}[\psi^{\mathrm{OI}}(X_0)g(X_0)].
\]
For $g=g_j$, it follows that
\[
    v_k^\top A_{t,N}^{1/2}
    \left.\frac{\mathrm{d}}{\mathrm{d}\varepsilon}
        c_\varepsilon^\star\right|_{\varepsilon=0}
    =\E_{\rho_0}[Z_kg_j]
    =\sqrt{\lambda_j}\delta_{kj}.
\]

Let $c_{\theta,n}^\star$ denote the population coefficient vector
corresponding to $\rho_{0,\theta,n}$. Since
$G_\theta=\sum_{j=1}^q\theta_jg_j$, the preceding derivative calculation
and smooth dependence of $A_\varepsilon^{-1}b_\varepsilon$ on
$\varepsilon$ give
\[
    v_j^\top A_{t,N}^{1/2}(c_{\theta,n}^\star-c^\star)
    =\varepsilon_n\theta_j\sqrt{\lambda_j}+O(\varepsilon_n^2),
\]
where the remainder is uniform in $\theta$ because $N$, and hence the
hypercube, is fixed. Therefore, for all sufficiently large $n$, the sign of
$v_j^\top A_{t,N}^{1/2}(c_{\theta,n}^\star-c^\star)$ is $\theta_j$, and its
absolute value is at least $\varepsilon_n\sqrt{\lambda_j}/2$.

Set $w_j:=\bigl(f_{A_{t,N}^{-1/2}v_j}\bigr)'$. By the definition of $A_{t,N}$, these
functions satisfy
$\langle f_u',f_v'\rangle_{L^2(\rho_t)}=u^\top A_{t,N}v$, and hence
\[
    \langle w_j,w_k\rangle_{L^2(\rho_t)}
        =v_j^\top A_{t,N}^{-1/2}A_{t,N}A_{t,N}^{-1/2}v_k
        =v_j^\top v_k
        =\delta_{jk}.
\]

Let $f_{t,N,\theta,n}^\star:=f_{c_{\theta,n}^\star}$ denote the population
OISM fit associated with the initial law $\rho_{0,\theta,n}$. By the
semigroup identity, this is also the population score-matching fit under
$\rho_{t,\theta,n}$. Hence, as in
Section~\ref{subsec:fixed-time-variance-growth},
\[
    (f_{t,N,\theta,n}^\star)'
    =\operatorname{proj}_{\mathcal P_{N-1}}^{L^2(\rho_{t,\theta,n})}
        s_{t,\theta,n}.
\]
Define
$\zeta_j(\theta):=\langle(f_{t,N,\theta,n}^\star)'-(f_{t,N}^\star)',
w_j\rangle_{L^2(\rho_t)}$. Then
$\zeta_j(\theta)=v_j^\top A_{t,N}^{1/2}(c_{\theta,n}^\star-c^\star)$, so for all
sufficiently large $n$ its sign is $\theta_j$ and
$|\zeta_j(\theta)|\geq\varepsilon_n\sqrt{\lambda_j}/2$.

Now let $\widehat s$ be an arbitrary measurable estimator based on the
initial sample. It is enough to consider estimators with finite risk under
every alternative, since otherwise the desired lower bound is immediate.
For large $n$, uniform equivalence of $\rho_{t,\theta,n}$ and $\rho_t$
ensures that finite $L^2(\rho_{t,\theta,n})$ risk gives the
$L^2(\rho_t)$ integrability required below. Set
$\widehat\zeta_j:=\langle\widehat s-(f_{t,N}^\star)',w_j
\rangle_{L^2(\rho_t)}$ and
$\widehat\theta_j:=\operatorname{sign}(\widehat\zeta_j)$, with any fixed
convention at zero. If $\widehat\theta_j\neq\theta_j$, then
$|\widehat\zeta_j-\zeta_j(\theta)|\geq|\zeta_j(\theta)|
\geq\varepsilon_n\sqrt{\lambda_j}/2$. Bessel's inequality therefore gives
\[
    \left\|\widehat s-(f_{t,N,\theta,n}^\star)'\right\|_{L^2(\rho_t)}^2
    \geq\frac{\varepsilon_n^2}{4}
        \sum_{j=1}^q\lambda_j
        1_{\{\widehat\theta_j\neq\theta_j\}}.
\]

It remains to show that neighboring vertices are statistically difficult to
distinguish. We use
\[
    \begin{aligned}
    \operatorname{TV}(P,Q)
        &:=\frac12\int|\mathrm{d}P-\mathrm{d}Q|,\\
    \chi^2(P\|Q)
        &:=\int\left(\frac{\mathrm{d}P}{\mathrm{d}Q}-1\right)^2
            \,\mathrm{d}Q,\qquad P\ll Q.
    \end{aligned}
\]
Let $\theta^{(j)}$ denote $\theta$ with its $j$th coordinate
flipped. Since $G_\theta-G_{\theta^{(j)}}=2\theta_jg_j$ and, for large $n$,
\enlargethispage{\baselineskip}
$1+\varepsilon_nG_{\theta^{(j)}}\geq1/2$, we have
$\chi^2(\rho_{0,\theta,n}\|\rho_{0,\theta^{(j)},n})
\leq16\varepsilon_n^2$. The inequality
$\operatorname{KL}(P\|Q)\leq\chi^2(P\|Q)$ and tensorization bound the
product KL divergence by $16n\varepsilon_n^2=16\delta^2$. Pinsker's
inequality therefore gives
\[
    \operatorname{TV}\!\left(
        \rho_{0,\theta,n}^{\otimes n},
        \rho_{0,\theta^{(j)},n}^{\otimes n}
    \right)
    \leq\sqrt{\frac12
        \operatorname{KL}\!\left(
            \rho_{0,\theta,n}^{\otimes n}
            \middle\|
            \rho_{0,\theta^{(j)},n}^{\otimes n}
        \right)}
    \leq\sqrt8\,\delta.
\]
Choose $\delta>0$ sufficiently small so that $\sqrt8\,\delta\leq1/2$.

We now apply the coordinatewise Assouad pairing argument
\citep{yu1997assouad}. For a fixed coordinate $j$, call
$\theta,\theta^{(j)}\in\{-1,1\}^q$ neighbors if they differ only in the
$j$th coordinate:
\[
    \theta_k^{(j)}=\theta_k\quad(k\neq j),
    \qquad
    \theta_j^{(j)}=-\theta_j.
\]
Pair each vertex $\theta$ with $\theta_j=1$ with its unique neighbor
$\theta^{(j)}$. The other $q-1$ coordinates are arbitrary, so there are
$2^{q-1}$ such disjoint pairs, and every vertex of $\{-1,1\}^q$ belongs
to exactly one pair. Since the correct $j$th signs under the two hypotheses
are $+1$ and $-1$, respectively, the sum of the two neighboring testing
error probabilities satisfies
\[
    \mathbb P_{\rho_{0,\theta,n}^{\otimes n}}
        (\widehat\theta_j=-1)
    +\mathbb P_{\rho_{0,\theta^{(j)},n}^{\otimes n}}
        (\widehat\theta_j=1)
    \geq1-\operatorname{TV}\!\left(
        \rho_{0,\theta,n}^{\otimes n},
        \rho_{0,\theta^{(j)},n}^{\otimes n}
    \right)
    \geq\frac12.
\]
Summing this inequality over the $2^{q-1}$ pairs gives
\[
    \sum_{\theta\in\{-1,1\}^q}
    \mathbb P_{\rho_{0,\theta,n}^{\otimes n}}
        (\widehat\theta_j\neq\theta_j)
    \geq2^{q-1}\cdot\frac12
    =2^{q-2},
\]
and hence
\[
    \frac1{2^q}\sum_{\theta\in\{-1,1\}^q}
    \mathbb P_{\rho_{0,\theta,n}^{\otimes n}}
        (\widehat\theta_j\neq\theta_j)
    \geq\frac14.
\]
Multiplying by $\lambda_j$, summing over $j$, and using that the supremum
over $\theta$ is at least the hypercube average gives
\[
    \begin{aligned}
    &\sup_\theta
        \E_{\rho_{0,\theta,n}^{\otimes n}}
        \left[
            \sum_{j=1}^q\lambda_j
            1_{\{\widehat\theta_j\neq\theta_j\}}
        \right]
    &\geq\frac1{2^q}
        \sum_{\theta\in\{-1,1\}^q}
        \E_{\rho_{0,\theta,n}^{\otimes n}}
        \left[
            \sum_{j=1}^q\lambda_j
            1_{\{\widehat\theta_j\neq\theta_j\}}
        \right]
        \geq\frac14\sum_{j=1}^q\lambda_j.
    \end{aligned}
\]
Since this inequality holds for the sign tests induced by every score
estimator $\widehat s$, combining it with the preceding Bessel bound gives
\[
    \inf_{\widehat s}\sup_\theta
    \E_{\rho_{0,\theta,n}^{\otimes n}}
    \left[
        \left\|\widehat s-(f_{t,N,\theta,n}^\star)'\right\|_{L^2(\rho_t)}^2
    \right]
    \geq\frac{\varepsilon_n^2}{16}
        \sum_{j=1}^q\lambda_j.
\]
This is the minimax risk measured in the fixed base norm $L^2(\rho_t)$.

To pass from the fixed base norm to the perturbed norm, the OU transition
kernel gives
\[
    \frac{\mathrm{d}\rho_{t,\theta,n}}{\mathrm{d}\rho_t}(X_t)
    =\E[1+\varepsilon_nG_\theta(X_0)\mid X_t].
\]
Hence
$1-\varepsilon_nK\leq
\mathrm{d}\rho_{t,\theta,n}/\mathrm{d}\rho_t
\leq1+\varepsilon_nK$. For all sufficiently large $n$, this implies
$\norm{f}_{L^2(\rho_{t,\theta,n})}^2
\geq\frac12\norm{f}_{L^2(\rho_t)}^2$ for every measurable $f$.
Combining the preceding estimates gives
\[
    \begin{aligned}
    &\inf_{\widehat s}\sup_\theta
    \E_{\rho_{0,\theta,n}^{\otimes n}}
    \left[
        \left\|
            \widehat s-
            \operatorname{proj}_{\mathcal P_{N-1}}^{L^2(\rho_{t,\theta,n})}
            s_{t,\theta,n}
        \right\|_{L^2(\rho_{t,\theta,n})}^2
    \right]
    &\geq\frac{\varepsilon_n^2}{32}
        \sum_{j=1}^q\lambda_j
    =\frac{\delta^2}{32n}V_N^{\mathrm{OI}}(t).
    \end{aligned}
\]
This proves the result.
\end{proof}

\section{Proofs for Fourier approximation under heat flow}
\label{app:heat-proofs}

This section proves the heat-flow results in two stages. We first separate
the integrated score error into approximation and estimation terms. Fourier
truncation controls the approximation term, while the identity (Lemma~\ref{lem:heat-energy})
controls the additional derivative in the Fourier bound after integration
in time. For the estimation term, a matrix lower bound imposed at time zero
propagates along the heat flow, and entropy dissipation controls the residual
covariance after integration in time. We then transfer the resulting score
bound to Kullback--Leibler loss and apply a classical lower bound for smooth
density estimation.

\subsection{Fourier coordinates and the population fit}
\label{subsec:heat-population-fit}

We use the Fourier coordinates and OISM coefficients
\eqref{eq:heat-fourier-coefficients} from Section~\ref{subsec:heat-score-error}.
For readability, throughout this appendix we suppress the superscript
$\mathrm{OI}$ and, for fixed $N$, the cutoff subscript on residuals and
covariances; the abbreviated symbols refer to these operator-informed
Fourier quantities. Write
$\widehat A_{t,N}:=\widehat A_{t,N}^{\mathrm{OI}}$ and
$\widehat b_{t,N}:=\widehat b_{t,N}^{\mathrm{OI}}$.
Define
\[
    c_{t,N}^\star:=A_{t,N}^{-1}b_{t,N},
    \qquad f_{t,N}^\star:=f_{c_{t,N}^\star}.
\]
On $\mathcal C_N$, write
$\widehat c_{t,N}:=\widehat c_{t,N}^{\mathrm{OI}}$. The population and empirical
minimizers are conjugate symmetric whenever they are uniquely defined, so
they represent real-valued potentials. Introduce the centered residual
\begin{equation}
    \psi_t(x):=\mathfrak b_t(x)-\mathfrak a_t(x)c_{t,N}^\star,
    \qquad
    \Sigma_t:=\E_{\rho_0}\!\left[
        \psi_t(X_0)\psi_t(X_0)^\dagger\right].
    \label{eq:heat-residual}
\end{equation}
The population normal equations give
$\E_{\rho_0}\psi_t(X_0)=0$.

\begin{lemma}[Population matrix bounds]
\label{lem:heat-population-matrix}
If $m\leq p_0\leq M$, then, for every $t\geq0$,
\begin{equation}
    mK_N\preceq A_{t,N}\preceq MK_N.
    \label{eq:heat-population-matrix-bounds}
\end{equation}
In particular, $A_{t,N}$ is positive definite.
\end{lemma}

\begin{proof}
The heat semigroup preserves the bounds $m\leq p_t\leq M$. For every
$c\in\mathbb C^{\Lambda_N}$, Parseval's identity gives
\[
    c^\dagger A_{t,N}c
        =\int_{\T^d}|\nabla f_c|^2p_t\,\mathrm{d}\nu,
    \qquad
    c^\dagger K_Nc
        =\int_{\T^d}|\nabla f_c|^2\,\mathrm{d}\nu.
\]
Multiplying the second identity by the lower and upper bounds for $p_t$
proves~\eqref{eq:heat-population-matrix-bounds}.
\end{proof}

To verify $|\Lambda_N|\asymp_d N^d$ in~\eqref{eq:heat-frequency-set},
note that the Euclidean ball of radius $N$ contains a cube of side length
$2N/\sqrt d$ and is contained in the cube $[-N,N]^d$. Counting lattice
points and removing the origin gives
\[
    (2\lfloor N/\sqrt d\rfloor+1)^d-1
        \leq |\Lambda_N|
        \leq (2\lfloor N\rfloor+1)^d-1.
\]
For $N\geq\sqrt d$, the inequality
$\lfloor N/\sqrt d\rfloor\geq N/(2\sqrt d)$ makes the lower bound
at least $d^{-d/2}N^d$; the upper bound is at most $3^dN^d$ for every
$N\geq1$. The same lower bound holds for $1\leq N<\sqrt d$ because
$\Lambda_N$ contains the coordinate unit vectors. Thus the comparison
constants depend only on $d$.

\subsection{Integrated approximation error}
\label{subsec:heat-integrated-approximation}

We first derive the derivative bounds supplied by the periodic H\"older
class. Fourier truncation of the score involves one additional derivative
of $\log p_t$, which the identity from
Lemma~\ref{lem:heat-energy} will control after integration in time.

\begin{lemma}[Derivative bounds for the periodic H\"older class]
\label{lem:heat-holder-derivatives}
Let $s\geq1$ be an integer. There exists a constant $C=C(d,s,L)$
such that, for every $p\in\mathcal P_s(m,M,L)$, $p$ has weak
derivatives through order $s$ and
\[
    \norm{D^\beta p}_{L^\infty(\nu)}\leq C,
    \qquad 1\leq|\beta|\leq s.
\]
For $|\beta|\leq s-1$, the derivatives are classical, while those of
order $s$ are understood weakly.
\end{lemma}

\begin{proof}
For $|\beta|=s-1$, the function $D^\beta p$ is $L$-Lipschitz.
By Rademacher's theorem and the weak differentiability of Lipschitz
functions, all weak derivatives of order $s$ exist and satisfy
\[
    \norm{D^\alpha p}_{L^\infty(\nu)}\leq L,
    \qquad |\alpha|=s.
\]
If $s=1$, this already proves the claim. We now assume $s\geq2$.

For $1\leq|\beta|\leq s-1$, periodicity gives
\[
    \int_{\T^d}D^\beta p\,\mathrm{d}\nu=0.
\]
Continuity and connectedness imply that $D^\beta p$ vanishes somewhere.
Since the torus has diameter $\pi\sqrt d$,
\[
    \norm{D^\beta p}_{L^\infty(\nu)}
        \leq\pi\sqrt d\,\operatorname{Lip}(D^\beta p).
\]
At order $s-1$, the defining Lipschitz bound for
$\mathcal P_s(m,M,L)$ therefore gives
\[
    \max_{|\beta|=s-1}\norm{D^\beta p}_\infty\leq\pi\sqrt d\,L.
\]
For $1\leq r\leq s-2$ and $|\beta|=r$,
\[
    \operatorname{Lip}(D^\beta p)
        \leq\norm{\nabla D^\beta p}_\infty
        \leq\sqrt d\max_{|\gamma|=r+1}\norm{D^\gamma p}_\infty.
\]
Combined with the zero-mean estimate above, this shows that a uniform
$L^\infty$ bound at order $r+1$ yields one at order $r$, and descending from
$r=s-2$ to $r=1$ proves the claim with $C=C(d,s,L)$ uniformly over
$p\in\mathcal P_s(m,M,L)$.
\end{proof}

This is the periodic analogue of the lower-derivative control in
Chewi et~al.\ \citep[Appendix~C.3]{chewi2026ddpm}.

\begin{corollary}[Derivative bounds along heat flow]
\label{cor:heat-propagated-regularity}
Let $p_0\in\mathcal P_s(m,M,L)$ and set $p_t=P_tp_0$ for $t\geq0$.
Then
\[
    m\leq p_t\leq M,\qquad t\geq0.
\]
Moreover, there exists $C=C(d,s,L)$ such that, for every multi-index
$\beta$ with $1\leq|\beta|\leq s$,
\[
    D^\beta p_t=P_t(D^\beta p_0),
    \qquad t\geq0.
\]
At $t=0$, derivatives of order $s$ are understood weakly; for every
$t>0$, all derivatives are classical. In particular,
\begin{equation}
    \sup_{t\geq0}\norm{D^\beta p_t}_{L^\infty(\nu)}
        \leq\norm{D^\beta p_0}_{L^\infty(\nu)}\leq C.
    \label{eq:heat-propagated-derivatives}
\end{equation}
For every $t>0$, both $p_t$ and $\log p_t$ belong to $C^\infty(\T^d)$.
\end{corollary}

\begin{proof}
Order preservation and $P_t1=1$ give $m\leq p_t\leq M$.
For $t>0$, $P_t$ is convolution with the smooth periodic heat kernel
$K_t$, so $P_tp_0\in C^\infty(\T^d)$ and its derivatives are classical,
with the following identity holding in the sense of distributions:
\[
    D^\beta P_tp_0=D^\beta(K_t*p_0)
        =K_t*D^\beta p_0=P_t(D^\beta p_0).
\]
Since $P_t$ is an
$L^\infty$-contraction, Lemma~\ref{lem:heat-holder-derivatives} yields
\[
    \norm{D^\beta p_t}_\infty
        \leq\norm{D^\beta p_0}_\infty\leq C.
\]
Since $p_t\geq m>0$, also $\log p_t\in C^\infty(\T^d)$ for $t>0$.
\end{proof}

\begin{lemma}[Fourier truncation of the score]
\label{lem:heat-fourier-approximation}
Fix an integer $s\geq1$ and $t>0$. If $0<p_t\leq M$ and
$u_t:=\log p_t\in C^{s+1}(\T^d)$, then
\begin{equation}
    B_N^{\mathrm{heat}}(t)
    \leq MN^{-2s}\sum_{i_0,\ldots,i_s=1}^d
        \norm{\partial_{i_0}\cdots\partial_{i_s}u_t}_{L^2(\nu)}^2.
    \label{eq:heat-pointwise-approximation}
\end{equation}
\end{lemma}

\begin{proof}
Define the Fourier coefficients and truncation
\[
    \widehat u_t(k):=\int_{\T^d}u_t(x)\overline{\omega_k(x)}\,\mathrm{d}\nu(x)
        =\int_{\T^d}u_t(x)e^{-i\langle k,x\rangle}\,\mathrm{d}\nu(x),
\]
\[
    \Pi_Nu_t:=\sum_{k\in\Lambda_N}\widehat u_t(k)\omega_k.
\]
Because $u_t$ is real-valued, $\Pi_Nu_t$ is an admissible potential.
The omitted constant has zero gradient, and Fourier truncation commutes
with differentiation. Using this potential in the approximation problem
and then applying $p_t\leq M$ and Parseval's identity gives
\[
    \begin{aligned}
    B_N^{\mathrm{heat}}(t)
        &\leq\norm{\nabla\Pi_Nu_t-\nabla u_t}_{L^2(\rho_t)}^2\\
        &\leq M\norm{\nabla\Pi_Nu_t-\nabla u_t}_{L^2(\nu)}^2\\
        &=M\sum_{|k|_2>N}|k|_2^2|\widehat u_t(k)|^2.
    \end{aligned}
\]
For $|k|_2>N$, we have $|k|_2^2\leq N^{-2s}|k|_2^{2(s+1)}$, so
    $B_N^{\mathrm{heat}}(t)
        \leq MN^{-2s}\sum_{k\in\mathbb Z^d}
            |k|_2^{2(s+1)}|\widehat u_t(k)|^2$.
The expansion
\[
    |k|_2^{2(s+1)}
        =\left(\sum_{j=1}^d k_j^2\right)^{s+1}
        =\sum_{i_0,\ldots,i_s=1}^d
            k_{i_0}^2\cdots k_{i_s}^2
\]
then gives
\[
    \sum_{k\in\mathbb Z^d}|k|_2^{2(s+1)}|\widehat u_t(k)|^2
        =\sum_{i_0,\ldots,i_s=1}^d\sum_{k\in\mathbb Z^d}
            k_{i_0}^2\cdots k_{i_s}^2|\widehat u_t(k)|^2.
\]
For each fixed tuple $(i_0,\ldots,i_s)$, the standard differentiation for Fourier coefficients gives
\[
    \widehat{\partial_{i_0}\cdots\partial_{i_s}u_t}(k)
        =(ik_{i_0})\cdots(ik_{i_s})\widehat u_t(k),
    \qquad
    \left|\widehat{\partial_{i_0}\cdots\partial_{i_s}u_t}(k)\right|^2
        =k_{i_0}^2\cdots k_{i_s}^2|\widehat u_t(k)|^2.
\]
Parseval's identity applied to this derivative therefore yields
\[
    \sum_{k\in\mathbb Z^d}k_{i_0}^2\cdots k_{i_s}^2|\widehat u_t(k)|^2
        =\norm{\partial_{i_0}\cdots\partial_{i_s}u_t}_{L^2(\nu)}^2,
\]
and summing over the tuples proves~\eqref{eq:heat-pointwise-approximation}.
\end{proof}

\begin{lemma}
\label{lem:heat-energy}
For every $h\in L^2(\nu)$ and $T>0$,
\begin{equation}
    \int_0^T\norm{\nabla P_th}_{L^2(\nu)}^2\,\mathrm{d}t
        =\frac12\left(
            \norm{h}_{L^2(\nu)}^2-
            \norm{P_Th}_{L^2(\nu)}^2
        \right)
        \leq\frac12\norm{h}_{L^2(\nu)}^2.
    \label{eq:heat-energy-identity}
\end{equation}
\end{lemma}

\begin{proof}
For smooth $h$, the heat equation and integration by parts give
\[
    \frac{\mathrm{d}}{\mathrm{d}t}\norm{P_th}_{L^2(\nu)}^2
        =2\langle P_th,\Delta P_th\rangle_{L^2(\nu)}
        =-2\norm{\nabla P_th}_{L^2(\nu)}^2.
\]
Integration over $[0,T]$ proves the identity. For general
$h\in L^2(\nu)$, apply the smooth calculation on $[\varepsilon,T]$ to obtain
\[
    \int_\varepsilon^T\norm{\nabla P_th}_{L^2(\nu)}^2\,\mathrm{d}t
        =\frac12\left(
            \norm{P_\varepsilon h}_{L^2(\nu)}^2
            -\norm{P_Th}_{L^2(\nu)}^2
        \right).
\]
By Parseval's identity and dominated convergence applied to the Fourier
coefficients, $P_\varepsilon h\to h$ in $L^2(\nu)$ as
$\varepsilon\downarrow0$, and therefore
$\norm{P_\varepsilon h}_{L^2(\nu)}^2\to\norm{h}_{L^2(\nu)}^2$.
Since the integrand is nonnegative and $[\varepsilon,T]$ increases to
$(0,T]$, monotone convergence gives
\[
    \int_\varepsilon^T\norm{\nabla P_th}_{L^2(\nu)}^2\,\mathrm{d}t
        \longrightarrow
    \int_0^T\norm{\nabla P_th}_{L^2(\nu)}^2\,\mathrm{d}t,
\]
proving~\eqref{eq:heat-energy-identity}.
\end{proof}

\begin{lemma}[Integrated regularity bound]
\label{lem:heat-integrated-approximation}
Let $p_0\in\mathcal P_s(m,M,L)$. There is
$C=C(d,s,m,M,L)$ such that, for every $T>0$,
\begin{equation}
    \sum_{i_0,\ldots,i_s=1}^d\int_0^T
        \norm{\partial_{i_0}\cdots\partial_{i_s}
            \log p_t}_{L^2(\nu)}^2\,\mathrm{d}t\leq C.
    \label{eq:heat-integrated-regularity}
\end{equation}
\end{lemma}

\begin{proof}
Corollary~\ref{cor:heat-propagated-regularity} gives all the regularity
needed below. For $t>0$, $p_t$ and $\log p_t$ are smooth, so the
classical multivariate Fa\`a di Bruno formula applies. Fix
$i_0,\ldots,i_s\in\{1,\ldots,d\}$ and, for a nonempty subset
$B\subseteq\{0,\ldots,s\}$, write
$\partial_{i_B}:=\prod_{\ell\in B}\partial_{i_\ell}$. Then
\begin{equation}
    \partial_{i_0}\cdots\partial_{i_s}\log p_t
    =\sum_{\mathcal Q}
        (-1)^{|\mathcal Q|-1}(|\mathcal Q|-1)!
        \prod_{B\in\mathcal Q}
            \frac{\partial_{i_B}p_t}{p_t},
    \label{eq:heat-log-faa-di-bruno}
\end{equation}
where $\mathcal Q$ ranges over the partitions of $\{0,\ldots,s\}$.

For each partition, we apply Lemma~\ref{lem:heat-energy} to the factor
containing the derivative $\partial_{i_0}$. Let $B_0$ be the unique
block containing the index $0$, and set
\[
    h_0:=
    \begin{cases}
        \partial_{i_{B_0\setminus\{0\}}}p_0,&B_0\neq\{0\},\\
        p_0,&B_0=\{0\}.
    \end{cases}
\]
Here the indicated derivative of $p_0$ is weak when necessary.
Its order is $|B_0|-1\leq s$, so
Lemma~\ref{lem:heat-holder-derivatives} gives $h_0\in L^2(\nu)$.
Commutation with the heat semigroup yields
\[
    \partial_{i_{B_0}}p_t=\partial_{i_0}P_th_0.
\]
This separates one derivative to which Lemma~\ref{lem:heat-energy} applies,
including when $B_0$ contains all $s+1$ indices.

Every other block $B\neq B_0$ excludes the index $0$ and therefore has
at most $s$ elements. Corollary~\ref{cor:heat-propagated-regularity}
and $p_t\geq m$ consequently give
\[
    \left\|\frac{\partial_{i_B}p_t}{p_t}\right\|_{L^\infty(\nu)}
        \leq C,\qquad B\neq B_0,
    \qquad
    \left\|\frac{\partial_{i_{B_0}}p_t}{p_t}\right\|_{L^2(\nu)}
        \leq m^{-1}\norm{\partial_{i_0}P_th_0}_{L^2(\nu)}.
\]
Bounding all but the distinguished factor in $L^\infty$ gives
\[
    \left\|\prod_{B\in\mathcal Q}
        \frac{\partial_{i_B}p_t}{p_t}\right\|_{L^2(\nu)}^2
        \leq C\norm{\partial_{i_0}P_th_0}_{L^2(\nu)}^2,
\]
where the contribution of the at most $s$ nondistinguished factors has been absorbed into the constant $C=C(d,s,m,M,L)$ (we treat the empty product as evaluating to $1$). Integrating and applying Lemma~\ref{lem:heat-energy}, we obtain
\begin{equation}
    \begin{aligned}
    \int_0^T\left\|\prod_{B\in\mathcal Q}
        \frac{\partial_{i_B}p_t}{p_t}\right\|_{L^2(\nu)}^2\,\mathrm{d}t
        &\leq C\int_0^T\norm{\nabla P_th_0}_{L^2(\nu)}^2\,\mathrm{d}t\\
        &\leq\frac C2\norm{h_0}_{L^2(\nu)}^2.
    \end{aligned}
    \label{eq:heat-selected-derivative-energy}
\end{equation}
The last norm is uniformly bounded. If $h_0=p_0$, then
$\norm{p_0}_{L^2(\nu)}^2\leq M\int p_0\,\mathrm{d}\nu=M$. Otherwise,
Lemma~\ref{lem:heat-holder-derivatives} bounds $h_0$ in $L^\infty(\nu)$,
and hence in $L^2(\nu)$ since $\nu$ is a probability measure.

Applying Cauchy--Schwarz to the finite sum over partitions in
\eqref{eq:heat-log-faa-di-bruno} gives
\[
    \begin{aligned}
    &\left\|\sum_{\mathcal Q}(-1)^{|\mathcal Q|-1}(|\mathcal Q|-1)!
        \prod_{B\in\mathcal Q}\frac{\partial_{i_B}p_t}{p_t}
        \right\|_{L^2(\nu)}^2\\
    &\qquad\leq
        \left(\sum_{\mathcal Q}\bigl((|\mathcal Q|-1)!\bigr)^2\right)
        \sum_{\mathcal Q}\left\|\prod_{B\in\mathcal Q}
            \frac{\partial_{i_B}p_t}{p_t}\right\|_{L^2(\nu)}^2.
    \end{aligned}
\]
Both sums run over the finitely many partitions of $\{0,\ldots,s\}$.
Their number and the coefficients $(|\mathcal Q|-1)!$ depend only on $s$,
while~\eqref{eq:heat-selected-derivative-energy} and the preceding bound
on $\norm{h_0}_{L^2(\nu)}$ uniformly bound the time integral of each term
in the second sum, so these finite factors can be absorbed into
$C=C(d,s,m,M,L)$. Therefore, for each fixed tuple $(i_0,\ldots,i_s)$,
\[
    \int_0^T\norm{\partial_{i_0}\cdots\partial_{i_s}
        \log p_t}_{L^2(\nu)}^2\,\mathrm{d}t\leq C,
\]
uniformly in $T$.
There are exactly $d^{s+1}$ such tuples in $\{1,\ldots,d\}^{s+1}$;
summing and absorbing this factor into $C$ proves
\eqref{eq:heat-integrated-regularity}.
\end{proof}

\begin{proof}[Proof of Proposition~\ref{prop:heat-integrated-approximation}]
For every $t>0$, the smoothness of $\log p_t$ permits the use of
Lemma~\ref{lem:heat-fourier-approximation}. Integrating its bound and
applying~\eqref{eq:heat-integrated-regularity} gives
\[
    \int_0^T B_N^{\mathrm{heat}}(t)\,\mathrm{d}t
        \leq MN^{-2s}\sum_{i_0,\ldots,i_s=1}^d
            \int_0^T\norm{\partial_{i_0}\cdots\partial_{i_s}
                \log p_t}_{L^2(\nu)}^2\,\mathrm{d}t
        \leq CN^{-2s},
\]
where the factor $M$ is absorbed into $C$.
\end{proof}

The pointwise Fourier bound involves $s+1$ derivatives of $\log p_t$,
but only $s$ derivatives of the initial density are required: after
integration in time, Lemma~\ref{lem:heat-energy} supplies the remaining
derivative.

\subsection{Variance and finite-sample estimation}
\label{subsec:heat-estimation-proof}

The empirical argument uses only one conditioning event. A lower bound
for the empirical matrix at time zero propagates to every later time by
the Schur product theorem. The estimation error can then be controlled by
the integrated covariance of the centered normal equations.

\begin{lemma}[Integrated residual covariance]
\label{lem:heat-integrated-residual}
If $p_0$ is a density bounded above by $M$, then
\begin{equation}
    \int_0^T\tr(K_N^{-1}\Sigma_t)\,\mathrm{d}t
        \leq|\Lambda_N|(1+2\log M).
    \label{eq:heat-integrated-residual}
\end{equation}
\end{lemma}

\begin{proof}
At positive times heat smoothing makes $p_t$ smooth and strictly positive,
so the population fit and its residual are well defined.
For $k\in\Lambda_N$, the first term of the residual satisfies
\[
    |(\mathfrak b_t(x))_k|^2
        =|k|_2^4e^{-2t|k|_2^2}.
\]
The second term has the conditional-expectation representation
\[
    (\mathfrak a_t(x)c_{t,N}^\star)_k
        =P_t\!\left(
            \overline{\nabla\omega_k}\cdot
            \nabla f_{t,N}^\star
        \right)(x).
\]
Conditional Jensen's inequality and the projection property give
\[
    \E_{\rho_0}|(\mathfrak a_t(X_0)c_{t,N}^\star)_k|^2
        \leq |k|_2^2
            \norm{\nabla f_{t,N}^\star}_{L^2(\rho_t)}^2
        \leq |k|_2^2\norm{s_t}_{L^2(\rho_t)}^2.
\]
Write $I(p_t):=\norm{s_t}_{L^2(\rho_t)}^2$ for the Fisher information.
Since $K_N=\operatorname{diag}(|k|_2^2:k\in\Lambda_N)$ and
$\Sigma_t=\E_{\rho_0}[\psi_t(X_0)\psi_t(X_0)^\dagger]$, the residual
$\psi_t=\mathfrak b_t-\mathfrak a_t c_{t,N}^\star$ and the coordinatewise
inequality $|a-b|^2\leq2|a|^2+2|b|^2$ give
\[
    \begin{aligned}
    \tr(K_N^{-1}\Sigma_t)
        &=\sum_{k\in\Lambda_N}
            \frac{\E_{\rho_0}|(\psi_t(X_0))_k|^2}{|k|_2^2}\\
        &\leq2\sum_{k\in\Lambda_N}
            \frac{\E_{\rho_0}|(\mathfrak b_t(X_0))_k|^2}{|k|_2^2}
            +2\sum_{k\in\Lambda_N}
            \frac{\E_{\rho_0}|(\mathfrak a_t(X_0)c_{t,N}^\star)_k|^2}{|k|_2^2}.
    \end{aligned}
\]
Inserting the preceding bounds cancels a factor $|k|_2^2$ in each term;
each quotient in the second sum is at most $I(p_t)$, yielding
\begin{equation}
    \tr(K_N^{-1}\Sigma_t)
        \leq2\sum_{k\in\Lambda_N}
            |k|_2^2e^{-2t|k|_2^2}
            +2|\Lambda_N|I(p_t).
    \label{eq:heat-pointwise-residual-bound}
\end{equation}
For each nonzero $k$,
\[
    \int_0^T|k|_2^2e^{-2t|k|_2^2}\,\mathrm{d}t\leq\frac12.
\]
Since $\partial_t p_t=\Delta p_t$, integration by parts on the torus gives,
for $t>0$,
\[
    \begin{aligned}
    \frac{\mathrm{d}}{\mathrm{d}t}\mathrm{KL}(p_t\|1)
        =\frac{\mathrm{d}}{\mathrm{d}t}\int_{\T^d}p_t\log p_t\,\mathrm{d}\nu
        =\int_{\T^d}(1+\log p_t)\Delta p_t\,\mathrm{d}\nu
        =-\int_{\T^d}\frac{|\nabla p_t|^2}{p_t}\,\mathrm{d}\nu
        =-I(p_t).
    \end{aligned}
\]
Therefore, by nonnegativity of $\mathrm{KL}(p_T\|1)$,
\[
    \int_0^T I(p_t)\,\mathrm{d}t
        \leq\mathrm{KL}(p_0\|1)\leq\log M.
\]
The last inequality uses $\log p_0\leq\log M$ wherever $p_0>0$,
with $0\log0=0$. For a nonsmooth $p_0$, apply this calculation first on
$[\varepsilon,T]$ and let $\varepsilon\downarrow0$; bounded convergence
handles the entropy terms and monotone convergence handles the Fisher
information. Integrating~\eqref{eq:heat-pointwise-residual-bound} and
combining these bounds gives
\[
    \begin{aligned}
    \int_0^T\tr(K_N^{-1}\Sigma_t)\,\mathrm{d}t
        &\leq2\sum_{k\in\Lambda_N}\int_0^T
            |k|_2^2e^{-2t|k|_2^2}\,\mathrm{d}t
            +2|\Lambda_N|\int_0^T I(p_t)\,\mathrm{d}t\\
        &\leq|\Lambda_N|+2|\Lambda_N|\log M\\
        &=|\Lambda_N|(1+2\log M),
    \end{aligned}
\]
which is~\eqref{eq:heat-integrated-residual}.
\end{proof}

\begin{lemma}[Empirical conditioning along heat flow]
\label{lem:heat-empirical-conditioning}
If $m\leq p_0$, then
\begin{equation}
    \mathbb P_{p_0}(\mathcal C_N^c)
        \leq|\Lambda_N|\exp\!\left\{
            -\frac{m(1-\log2)n}{2|\Lambda_N|}
        \right\}.
    \label{eq:heat-conditioning-probability}
\end{equation}
Moreover, on $\mathcal C_N$,
\begin{equation}
    \widehat A_{t,N}\succeq\frac m2K_N
    \qquad\text{for every }t\geq0.
    \label{eq:heat-conditioning-propagation}
\end{equation}
\end{lemma}

\begin{proof}
Set
\[
    \widetilde{\mathfrak a}_0(x)
        :=K_N^{-1/2}\mathfrak a_0(x)K_N^{-1/2},
    \qquad
    G_0:=\E_{\rho_0}\widetilde{\mathfrak a}_0(X_0)=K_N^{-1/2}A_{0,N}K_N^{-1/2}.
\]
These matrices are positive semidefinite,
$G_0\succeq mI$ by the population lower bound, so $G_0^{-1}\preceq m^{-1}I$.
Also,
\[
    \norm{\widetilde{\mathfrak a}_0(x)}_{\mathrm{op}}
        \leq\tr\widetilde{\mathfrak a}_0(x)=|\Lambda_N|.
\]
Whiten by $G_0$ and define the independent matrices
\[
    Y_i:=G_0^{-1/2}\widetilde{\mathfrak a}_0(X_0^{(i)})G_0^{-1/2}.
\]
They satisfy $\E Y_i=I$ and
\[
    0\preceq Y_i\preceq\frac{|\Lambda_N|}{m}I.
\]
The matrix Chernoff inequality
\citep[Corollary~5.2]{tropp2012tail}, with lower-tail parameter $1/2$, gives
\begin{equation}
    \mathbb P_{p_0}\!\left(\widehat A_{0,N}\not\succeq\frac12A_{0,N}\right)
        \leq|\Lambda_N|\exp\!\left\{
            -\frac{m(1-\log2)n}{2|\Lambda_N|}
        \right\}.
    \label{eq:heat-matrix-chernoff}
\end{equation}
Because $A_{0,N}\succeq mK_N$, the event
$\{\widehat A_{0,N}\succeq A_{0,N}/2\}$ is contained in $\mathcal C_N$.
This proves~\eqref{eq:heat-conditioning-probability}.

For propagation, write
\[
    \mathfrak a_t(x)=\mathfrak a_0(x)\circ C_t,
    \qquad
    (C_t)_{k\ell}:=e^{-t|k-\ell|_2^2},
\]
where $\circ$ is the Hadamard product. The Gaussian-kernel matrix
$C_t$ is positive semidefinite and $K_N\circ C_t=K_N$. On
$\mathcal C_N$, the Schur product theorem therefore gives
\[
    \widehat A_{t,N}-\frac m2K_N
        =\left(\widehat A_{0,N}-\frac m2K_N\right)\circ C_t
        \succeq0,
\]
which proves~\eqref{eq:heat-conditioning-propagation}.
\end{proof}

\begin{proof}[Proof of Proposition~\ref{prop:heat-variance-estimation}]
In the complex Fourier coordinates used here, the asymptotic variance
is $V_N^{\mathrm{OI}}(t)=\tr(A_{t,N}^{-1}\Sigma_t)$. Since
$A_{t,N}\succeq mK_N$, we have $A_{t,N}^{-1}\preceq m^{-1}K_N^{-1}$.
Together with $\Sigma_t\succeq0$,
Lemma~\ref{lem:heat-integrated-residual} gives
\[
    \int_0^T V_N^{\mathrm{OI}}(t)\,\mathrm{d}t
        \leq\frac1m\int_0^T\tr(K_N^{-1}\Sigma_t)\,\mathrm{d}t
        \leq\frac{|\Lambda_N|}{m}(1+2\log M),
\]
which is~\eqref{eq:heat-integrated-variance-bound}.

For the finite-sample estimation bound, let $\overline\psi_t:=\frac1n\sum_{i=1}^n\psi_t(X_0^{(i)}).$
On $\mathcal C_N$, the empirical normal equations give
    $\widehat A_{t,N}(\widehat c_{t,N}-c_{t,N}^\star)
        =\overline\psi_t$.
Let $e_t:=\widehat c_{t,N}-c_{t,N}^\star$. The empirical lower bound
$\widehat A_{t,N}\succeq(m/2)K_N$ and the normal equations imply
\[
    \norm{K_N^{1/2}e_t}_2
        \leq\frac2m\norm{K_N^{-1/2}\overline\psi_t}_2.
\]
Combining this with the population upper bound in
Lemma~\ref{lem:heat-population-matrix} gives
\[
    \begin{aligned}
    \norm{\widehat s_{t,N}-\nabla f_{t,N}^\star}_{L^2(\rho_t)}^2
        =e_t^\dagger A_{t,N}e_t
        \leq M\norm{K_N^{1/2}e_t}_2^2
        \leq\frac{4M}{m^2}
            \norm{K_N^{-1/2}\overline\psi_t}_2^2.
    \end{aligned}
\]
Independence and centering of the residuals imply
\[
    \E_{p_0}\norm{K_N^{-1/2}\overline\psi_t}_2^2
        =\frac1n\tr(K_N^{-1}\Sigma_t).
\]
Lemma~\ref{lem:heat-integrated-residual} and Tonelli's theorem therefore
prove~\eqref{eq:heat-expected-estimation-bound}. The conditioning
probability in the proposition is
Lemma~\ref{lem:heat-empirical-conditioning}.
\end{proof}

\begin{proof}[Proof of Theorem~\ref{thm:heat-expected-score-risk}]
The projection identity gives, on $\mathcal C_N$,
\[
    \norm{\widehat s_{t,N}-s_t}_{L^2(\rho_t)}^2
        =\norm{\widehat s_{t,N}-\nabla f_{t,N}^\star}_{L^2(\rho_t)}^2
            +B_N^{\mathrm{heat}}(t).
\]
On $\mathcal C_N^c$, the estimator is zero, so
\[
    \mathcal R_T(0)=\int_0^T I(p_t)\,\mathrm{d}t\leq\log M.
\]
Propositions~\ref{prop:heat-integrated-approximation}
and~\ref{prop:heat-variance-estimation}, together with
Lemma~\ref{lem:heat-empirical-conditioning} yield
\begin{equation}
    \begin{aligned}
    \E_{p_0}\mathcal R_T(\widehat s_{\cdot,N})
        \leq{}&CN^{-2s}\\
        &+\frac{4M|\Lambda_N|}{m^2n}(1+2\log M)\\
        &+(\log M)|\Lambda_N|\exp\!\left\{
            -\frac{m(1-\log2)n}{2|\Lambda_N|}
        \right\}.
    \end{aligned}
    \label{eq:heat-expected-score-bound-explicit}
\end{equation}
Here $C=C(d,s,m,M,L)$ is independent of $T$. This
proves~\eqref{eq:heat-expected-score-bound} with constants uniform in $T$.

If $N_n\asymp n^{1/(2s+d)}$, then
$N_n^{-2s}\asymp|\Lambda_{N_n}|/n\asymp
n^{-2s/(2s+d)}$, while the last term in
\eqref{eq:heat-expected-score-bound-explicit} is exponentially smaller.
This proves~\eqref{eq:heat-expected-score-rate}.
\end{proof}

\subsection{Kullback--Leibler risk and the minimax lower bound}
\label{subsec:heat-kl-minimax-proof}

It remains to connect the integrated score error with the learned reverse
diffusion. The upper bound follows from a path-space change of measure and
the exponential convergence of the forward heat flow to the uniform
distribution $\nu$. The
matching lower bound follows by localizing the classical H\"older
density-estimation lower bound to the torus and then applying Pinsker's
inequality.

\begin{proof}[Proof of \eqref{eq:heat-kl-transfer}]
Fix the training data. Lemma~\ref{lem:heat-holder-derivatives} and
Corollary~\ref{cor:heat-propagated-regularity} give
\[
    \sup_{t\geq0}\norm{\nabla p_t}_\infty\leq C,
    \qquad p_t\geq m.
\]
At $t=0$, the gradient is interpreted weakly when $s=1$. The weak
chain rule for the logarithm, which has bounded derivative on
$[m,M]$, then yields
\begin{equation}
    \sup_{t\geq0}\norm{s_t}_\infty\leq\frac Cm,
    \label{eq:heat-uniform-score-bound}
\end{equation}
with the same weak interpretation of $s_0$.
On $\mathcal C_N$, equation~\eqref{eq:heat-conditioning-propagation},
together with continuity of $t\mapsto\widehat A_{t,N}$ and
$t\mapsto\widehat b_{t,N}$, makes
$t\mapsto\widehat c_{t,N}=\widehat A_{t,N}^{-1}\widehat b_{t,N}$ continuous.
Since the Fourier space is finite-dimensional, the map
$(t,x)\mapsto\widehat s_{t,N}(x)$ is continuous and therefore bounded on
the compact set $[0,T]\times\T^d$. On
$\mathcal C_N^c$, it is zero. Since the fitted score is a finite Fourier
sum with coefficients continuous in time, its spatial derivatives are
uniformly bounded on this compact time interval. Hence the learned drift
is globally Lipschitz in space, uniformly in time. The learned reverse SDE is therefore well
defined.

Fix $\varepsilon\in(0,T)$. Let $\mathbb Q^\varepsilon$ be the path law
on $[0,T-\varepsilon]$ of
\[
    \mathrm{d}X_t^{\leftarrow}
        =2s_{T-t}(X_t^{\leftarrow})\,\mathrm{d}t+\sqrt2\,\mathrm{d}B_t,
    \qquad X_0^{\leftarrow}\sim\rho_T,
\]
and let $\mathbb P^\varepsilon$ be the conditional path law of the
approximate reverse process $\widehat X^{\leftarrow}$ on the same interval,
initialized from $\nu$. Following the path-space formulation of
Chen et~al.\ \citep[\S5.1]{chen2023sampling}, these are measures on the same space of
continuous torus-valued paths, under which the coordinate process has the
exact and learned reverse drifts, respectively. The terminal marginal of
$\mathbb Q^\varepsilon$ is $p_\varepsilon\,\mathrm{d}\nu$. The truncation ensures
$T-t\in[\varepsilon,T]$, where heat smoothing gives classical regularity.

The exact and learned reverse drifts are $2s_{T-t}$ and
$2\widehat s_{T-t,N}$, respectively, with common diffusion coefficient
$\sqrt2\,I$. The process appearing in Novikov's condition is obtained by
evaluating
\[
    (\sqrt2\,I)^{-1}
        \bigl(2s_{T-t}(x)-2\widehat s_{T-t,N}(x)\bigr)
        =\sqrt2\,\bigl(s_{T-t}(x)-\widehat s_{T-t,N}(x)\bigr)
\]
along $X_t^{\leftarrow}$.
By~\eqref{eq:heat-uniform-score-bound} and the fitted-score bound above,
conditional on the training data,
\[
    \sup_{\substack{0\leq t\leq T-\varepsilon\\x\in\T^d}}
        |s_{T-t}(x)-\widehat s_{T-t,N}(x)|_2<\infty.
\]
Since
\[
    \frac12\norm{(\sqrt2\,I)^{-1}
        \bigl(2s_{T-t}-2\widehat s_{T-t,N}\bigr)}_2^2
        =\norm{s_{T-t}-\widehat s_{T-t,N}}_2^2,
\]
the exponent in Novikov's condition is bounded by a finite deterministic
quantity conditional on the training data, and hence
\[
    \E_{\mathbb Q^\varepsilon}\exp\!\left\{
        \int_0^{T-\varepsilon}
            \norm{s_{T-t}(X_t^{\leftarrow})
                -\widehat s_{T-t,N}(X_t^{\leftarrow})}_2^2\,\mathrm{d}t\right\}
        <\infty.
\]
Thus, in this compact setting, Novikov's condition holds directly.
The factor $1/4=\frac12(\sqrt2)^{-2}$ below comes from the common
diffusion coefficient; the entropy chain rule and Girsanov's theorem give
\begin{equation}
    \begin{aligned}
    \mathrm{KL}(\mathbb Q^\varepsilon\|\mathbb P^\varepsilon)
        &=\mathrm{KL}(p_T\|1)
          +\frac14\E_{\mathbb Q^\varepsilon}
            \int_0^{T-\varepsilon}
            \norm{2s_{T-t}(X_t^{\leftarrow})
                -2\widehat s_{T-t,N}(X_t^{\leftarrow})}_2^2\,\mathrm{d}t\\
        &=\mathrm{KL}(p_T\|1)
          +\int_\varepsilon^T
            \norm{\widehat s_{r,N}-s_r}_{L^2(\rho_r)}^2\,\mathrm{d}r.
    \end{aligned}
    \label{eq:heat-path-kl}
\end{equation}
Since the initial laws of $\mathbb Q^\varepsilon$ and $\mathbb P^\varepsilon$
are $\rho_T$ and $\nu$, respectively, and $p_T=\mathrm{d}\rho_T/\mathrm{d}\nu$, the initial
contribution is $\mathrm{KL}(\rho_T\|\nu)=\mathrm{KL}(p_T\|1)$.
Under $\mathbb Q^\varepsilon$, $X_t^{\leftarrow}\sim\rho_{T-t}$, so the
drift contribution is the time integral of the squared $L^2(\rho_{T-t})$
score error, and the change of variables $r=T-t$ gives the second equality
in~\eqref{eq:heat-path-kl}.

Data processing for the terminal-time evaluation map bounds the relative
entropy between the terminal marginals by the right-hand side of
\eqref{eq:heat-path-kl}. As $\varepsilon\downarrow0$,
$p_\varepsilon\to p_0$ in $L^1(\nu)$, hence
$\rho_\varepsilon\Rightarrow\rho_0$, while continuity of the learned
diffusion paths gives weak convergence of its law at
$T-\varepsilon$ to its law at $T$. Joint lower semicontinuity of relative
entropy therefore gives
\[
    \mathrm{KL}\!\left(
        \rho_0\,\middle\|\,\operatorname{Law}(\widehat X_T^{\leftarrow})
    \right)
    \leq\liminf_{\varepsilon\downarrow0}
    \mathrm{KL}\!\left(
        \rho_\varepsilon\,\middle\|\,
        \operatorname{Law}(\widehat X_{T-\varepsilon}^{\leftarrow})
    \right).
\]
Combining this with the preceding data-processing bound and monotone
convergence of the nonnegative score-error integrals over
$[\varepsilon,T]\uparrow(0,T]$ proves
\eqref{eq:heat-kl-transfer}.
\end{proof}

The forward initialization error satisfies
\begin{equation}
    \begin{aligned}
    \mathrm{KL}(p_T\|1)
        &\leq\norm{p_T-1}_{L^2(\nu)}^2\\
        &\leq e^{-2T}\norm{p_0-1}_{L^2(\nu)}^2
        \leq(M-1)e^{-2T}.
    \end{aligned}
    \label{eq:heat-kl-initialization}
\end{equation}
The first inequality follows from $\log x\leq x-1$, the second from
Parseval's identity and the contraction factor $e^{-t|k|_2^2}\leq e^{-t}$
at every nonzero frequency, and the last from
$\int p_0^2\,\mathrm{d}\nu\leq M\int p_0\,\mathrm{d}\nu=M$.

\begin{proof}[Proof of Corollary~\ref{cor:heat-density-kl}]
Taking expectations in~\eqref{eq:heat-kl-transfer} and applying
Theorem~\ref{thm:heat-expected-score-risk} and
\eqref{eq:heat-kl-initialization}, with $|\Lambda_N|\asymp_d N^d$, gives
\[
    \sup_{p_0\in\mathcal P_s(m,M,L)}
        \E_{p_0}\mathrm{KL}(p_0\|\widehat p_{N,T})
    \leq C\left(
        e^{-2T}+N^{-2s}+\frac{N^d}{n}
        +N^de^{-cn/N^d}
    \right).
\]
Here $C=C(d,s,m,M,L)$ and $c=c(d,m)>0$.
For the choices $N=N_n$ and $T=T_n$ in the corollary,
\[
    e^{-2T_n}=n^{-2s/(2s+d)},
    \qquad
    N_n^{-2s}\asymp\frac{|\Lambda_{N_n}|}{n}
        \asymp n^{-2s/(2s+d)},
\]
and the exponential term is smaller. This proves
\eqref{eq:heat-density-kl-upper}.

For the matching lower bound, we give an explicit construction for
$\mathcal P_s(m,M,L)$ using the standard Assouad hypercube argument.
See Yang and Barron~\citep{yang1999information} for the classical information-theoretic
minimax framework and smooth density-estimation rates.
Identify $\T^d$ with
\[
    F:=(-\pi,\pi]^d,\qquad \mathrm{d}\nu(x)=(2\pi)^{-d}\,\mathrm{d}x,\quad x\in F,
\]
and choose
\[
    Q:=(-1,1)^d,\qquad \overline Q=[-1,1]^d\subset(-\pi,\pi)^d.
\]
Fix a nonzero $\varphi\in C_c^\infty((-1/2,1/2)^d)$ with
$\int_{\R^d}\varphi(x)\,\mathrm{d}x=0$. For sufficiently small $h>0$, a regular
grid packing inside $Q$ gives centers $x_1,\ldots,x_{J_h}$ such that
$x_j+h\operatorname{supp}\varphi$ are pairwise disjoint and lie compactly
inside $Q$, with $J_h\asymp h^{-d}$. Set
\[
    \varphi_{j,h}(x):=h^s\varphi\!\left(\frac{x-x_j}{h}\right),
    \qquad
    p_\theta(x):=1+a\sum_{j=1}^{J_h}\theta_j\varphi_{j,h}(x),
    \qquad \theta\in\{-1,1\}^{J_h},
\]
on $F$, and extend $p_\theta$ $2\pi\mathbb Z^d$-periodically.
It equals $1$ near the boundary of $F$, so this extension is smooth
across opposite faces.

We choose $a>0$ sufficiently small, fixed independently of $h$.
Every bump has zero integral, so $\int_{\T^d}p_\theta\,\mathrm{d}\nu=1$.
Disjoint supports give
\[
    \norm{\sum_j\theta_j\varphi_{j,h}}_\infty
        \leq h^s\norm{\varphi}_\infty.
\]
For $h\leq1$, a sufficiently small $a$ therefore ensures
$m\leq p_\theta\leq M$. For $|\beta|=s-1$,
\[
    D^\beta\varphi_{j,h}(x)
        =hD^\beta\varphi\!\left(\frac{x-x_j}{h}\right),
    \qquad
    \operatorname{Lip}(D^\beta\varphi_{j,h})\leq C_\varphi,
\]
uniformly in $h$. Indeed, the gradient of the displayed derivative
is bounded by a constant depending only on the fixed bump.
Since the supports are disjoint and the zero extensions are smooth,
the same gradient bound holds for the sum and its periodic extension.
Taking $a$ sufficiently small thus ensures
$\max_{|\beta|=s-1}\operatorname{Lip}(D^\beta p_\theta)\leq L$.
Consequently, every $p_\theta$ belongs to $\mathcal P_s(m,M,L)$,
with $a$ depending only on $d,s,m,M,L$ and the fixed bump.

Let $\theta^{(j)}$ denote $\theta$ with only its $j$th sign flipped.
The neighboring $L^1(\nu)$ separation is
\[
    \norm{p_\theta-p_{\theta^{(j)}}}_{L^1(\nu)}
        =2a(2\pi)^{-d}h^{s+d}\norm{\varphi}_{L^1(\R^d)}
        \asymp h^{s+d}.
\]
Disjoint supports make the $L^1$ distance between any two members
equal to this separation times the number of differing signs.
Since $p_{\theta^{(j)}}\geq m$, the neighboring one-sample KL divergence
satisfies
\[
    \mathrm{KL}(p_\theta\|p_{\theta^{(j)}})
        \leq\frac1m\norm{p_\theta-p_{\theta^{(j)}}}_{L^2(\nu)}^2
        =\frac{4a^2}{m(2\pi)^d}h^{2s+d}\norm{\varphi}_{L^2(\R^d)}^2
        \leq Ch^{2s+d}.
\]
For $n$ independent observations, additivity of relative entropy gives
\[
    \mathrm{KL}\!\left(
        (p_\theta\,\mathrm{d}\nu)^{\otimes n}\,\middle\|\,
        (p_{\theta^{(j)}}\,\mathrm{d}\nu)^{\otimes n}
    \right)\leq Cnh^{2s+d}.
\]
Choose $h\asymp n^{-1/(2s+d)}$ with the comparison constant sufficiently
small. Using the same total-variation convention as above, write
$\operatorname{TV}(p,q):=\operatorname{TV}(p\,\mathrm{d}\nu,q\,\mathrm{d}\nu)$
for densities.
Pinsker's inequality therefore bounds the total variation distance
between neighboring product measures by $1/2$. Assouad's lemma
\citep{yu1997assouad} then yields
\[
    \inf_{\widetilde p_n}\sup_{\theta\in\{-1,1\}^{J_h}}
        \E_{p_\theta}\operatorname{TV}(\widetilde p_n,p_\theta)
        \geq cJ_hh^{s+d}
        \asymp h^s
        \asymp n^{-s/(2s+d)}.
\]
Since this family lies in $\mathcal P_s(m,M,L)$, it follows that,
for some $c_0=c_0(d,s,m,M,L)>0$ and all sufficiently large $n$,
\[
    \inf_{\widetilde p_n}
    \sup_{p\in\mathcal P_s(m,M,L)}
        \E_p\operatorname{TV}(\widetilde p_n,p)
        \geq c_0 n^{-s/(2s+d)}.
\]
The infimum is over all measurable density-valued estimators based on
$n$ independent observations. No bounds are imposed on their output
densities.

To pass to KL loss, Pinsker's inequality gives
$\mathrm{KL}(p\|\widetilde p_n)\geq
2\operatorname{TV}(p,\widetilde p_n)^2$. Therefore Jensen's
inequality yields, for every such estimator,
\[
    \begin{aligned}
    \sup_{p\in\mathcal P_s(m,M,L)}
        \E_p\mathrm{KL}(p\|\widetilde p_n)
    &\geq2\sup_{p\in\mathcal P_s(m,M,L)}
        \E_p\operatorname{TV}(p,\widetilde p_n)^2\\
    &\geq2\left(
        \sup_{p\in\mathcal P_s(m,M,L)}
        \E_p\operatorname{TV}(p,\widetilde p_n)
        \right)^2.
    \end{aligned}
\]
Taking the infimum and applying the total variation lower bound gives
\[
    \inf_{\widetilde p_n}
    \sup_{p\in\mathcal P_s(m,M,L)}
        \E_p\mathrm{KL}(p\|\widetilde p_n)
        \geq2c_0^2n^{-2s/(2s+d)}.
\]
This matches~\eqref{eq:heat-density-kl-upper} and proves minimax
optimality.
\end{proof}

\end{document}